\documentclass{article} % For LaTeX2e
\usepackage{iclr2023_conference,times}

\usepackage{amsmath,amsfonts,bm}

\def\eqref#1{equation~\ref{#1}}
\def\1{\bm{1}}

\DeclareMathAlphabet{\mathsfit}{\encodingdefault}{\sfdefault}{m}{sl}
\SetMathAlphabet{\mathsfit}{bold}{\encodingdefault}{\sfdefault}{bx}{n}

\definecolor{mycitecolor}{rgb}{0, 0.4, 0.7}
\usepackage[breaklinks=true,bookmarks=true,citecolor=mycitecolor,colorlinks,pagebackref=true]{hyperref}
\usepackage{url}

\usepackage{multirow}
\usepackage{multicol}
\usepackage{caption}
\usepackage{mathtools}
\usepackage{graphicx}
\usepackage{amsmath}
\usepackage{amssymb}
\usepackage{booktabs}
\usepackage{enumitem}

\usepackage{floatrow}
\newfloatcommand{capbtabbox}{table}[][\FBwidth]

\newcommand*\samethanks[1][\value{footnote}]{\footnotemark[#1]}

\title{CLIP-ViP: Adapting Pre-trained Image-Text Model to Video-Language Alignment}

\author{Hongwei Xue$^1$\thanks{Equal contributon. This work was performed when Hongwei Xue and Yuchong Sun were visiting Microsoft Research as research interns.}, Yuchong Sun$^2$\samethanks[1], Bei Liu$^3$\footnotemark[2], Jianlong Fu$^3$\footnotemark[2], \\ {\bf Ruihua Song$^2$, Houqiang Li$^1$, Jiebo Luo$^4$} \\
$^1$University of Science and Technology of China, Hefei, China, \\
$^2$Renmin University of China, Beijing, China,\\
$^3$Microsoft Research, Beijing, China,\\
$^4$University of Rochester, Rochester, NY
}

\iclrfinalcopy % Uncomment for camera-ready version, but NOT for submission.
\begin{document}

\maketitle
\renewcommand{\thefootnote}{\fnsymbol{footnote}}
\footnotetext[2]{Corresponding authors.}

\newcommand{\tabpre}{
\begin{table}[t]
    \scriptsize
    \centering
    \begin{tabular}{l c c c c} 
    \toprule
    P-PT Data & R@1 $\uparrow$ & R@5 $\uparrow$ & R@10 $\uparrow$ & Mean $\uparrow$ \\
    \cmidrule{1-5}
    None  & 43.4 & 70.9 & 81.1 & 65.1 \\
    WV-2.5M \citep{bain2021frozen}& 43.1 & 71.0 &  80.8 & 65.0\\ 
    HV-100M \citep{xue2022advancing}& 44.7 & 71.3 &  81.2 & 65.7\\
    HV-10M & -
    & - & - & -\\
   
    \bottomrule
    \end{tabular}
    \caption{MSR-VTT~\citep{xu2016msr} text-to-video retrieval results of fine-tuning CLIP-meanPool post-pretrained on WebVid-2.5M (WV-2.5M), HD-VILA-100M (HV-100M), and a $10\%$ random sampled subset HV-10M. Mean $\uparrow$ indicates average value of Recall@1, 5 and 10.}
    \label{tab:analysis_data}
\end{table}
}

\newcommand{\figprecurve}{
\begin{figure}
    \centering
    \includegraphics[width=0.4\linewidth]{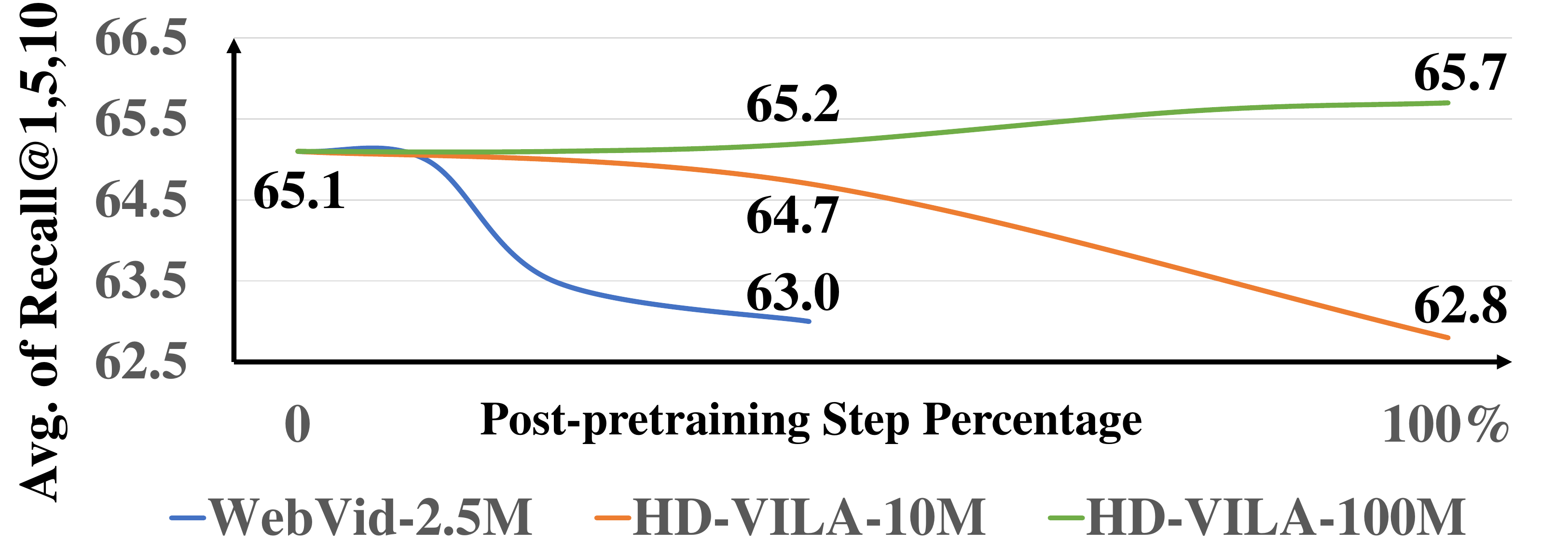}
    \caption{The curve of finetuning results during post-pretraining. The x-axis indicates the percentage of training steps. The y-axis indicates the average value of Recall@1, 5, and 10. [Best viewed in color]}
    \label{fig:precurve}
\end{figure}
}

\newcommand{\figmerge}{
\begin{figure}
\begin{floatrow}
\ffigbox{%
  \includegraphics[width=\linewidth]{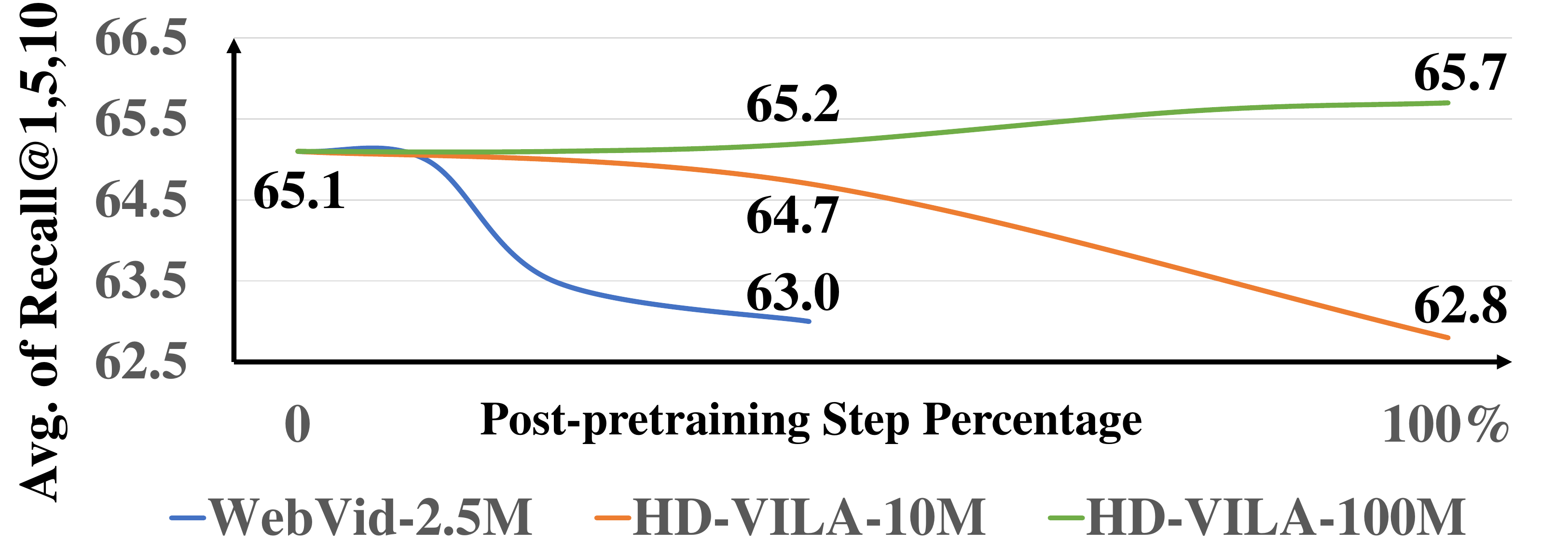}
}
{%
  \caption{The curve of finetuning results during post-pretraining. The x-axis indicates the percentage of training steps. The y-axis indicates average value of Recall@1, 5 and 10. [Best viewed in color]
  \label{fig:precurve}}%
}
\capbtabbox{%
    \scriptsize
  \begin{tabular}{l c c c } 
    \toprule
    NMI Score & MSR-VTT & DiDeMo  &  Mean\\
    \midrule
    HD-VILA$_{sub}$  & 0.831 & 0.684 & 0.758  \\
    HD-VILA$_{cap}$  & 0.317 & 0.621 & 0.469 \\
    WebVid  & 0.420 & 0.488 & 0.454 \\
    COCO  & 0.373 & 0.758 & 0.566 \\
    CC12M  & 0.445 & 0.673 & 0.559 \\
    \bottomrule
    \end{tabular}
}{%
  \caption{Normalized Mutual Information (NMI) score of language features extracted on series of data and downstream tasks. Larger value indicates larger domain gap.}%
  \label{tab:analysis_nmi}
}
\end{floatrow}
\end{figure}
}

\newcommand{\tabprenmi}{
\begin{table}[t]
    \scriptsize
    \centering
    \begin{tabular}{l c c c } 
    \toprule
    NMI Score & MSR-VTT & DiDeMo  &  Mean\\
    \midrule
    HD-VILA$_{sub}$  & 0.831 & 0.684 & 0.758  \\
    HD-VILA$_{cap}$  & 0.317 & 0.621 & 0.469 \\
    WebVid \citep{bain2021frozen} & 0.420 & 0.488 & 0.454 \\
    COCO \citep{lin2014microsoft} & 0.373 & 0.758 & 0.566 \\
    CC12M \citep{sharma2018conceptual}    & 0.445 & 0.673 & 0.559 \\
    \bottomrule
    \end{tabular}
    \caption{Normalized Mutual Information (NMI) score of language features extracted on series of data and downstream tasks. We choose MSR-VTT \citep{xu2016msr} and DiDeMo \citep{anne2017localizing} as downstream tasks. Larger value indicates larger domain gap.}
    \label{tab:analysis_nmi}
\end{table}
}

\newcommand{\tabablmodel}{
\begin{table}[t]
    \scriptsize
    \centering
    \begin{tabular}{l c c c c} 
    \toprule
    Model & R@1 $\uparrow$ & R@5 $\uparrow$ & R@10 $\uparrow$ & Mean $\uparrow$ \\
    \cmidrule{1-5}
    MeanPool  & 43.4 & 70.9 & 81.1 & 65.1 \\
    SeqTransformer & 44.6 & 71.2 & 81.8 & 65.9 \\
    Full Attention & 42.8 & 70.1 & 80.3 & 64.4 \\ 
    2 Video Proxy Tokens & 45.8 & 71.3 &  81.7 & 66.3 \\ 
    4 Video Proxy Tokens & \bf46.5 & 72.1 & \bf82.5 & \bf67.0 \\
    8 Video Proxy Tokens & 45.7 & \bf72.7 &  81.7 & 66.7 \\
    \bottomrule
    \end{tabular}
    \caption{MSR-VTT text-to-video retrieval results of finetuning CLIP by different settings. Mean $\uparrow$ indicates the average value of Recall@1, 5, and 10. All results are based on CLIP-ViT-B/32.}
    \label{tab:model}
\end{table}
}

\newcommand{\tabablloss}{
\begin{table*}[t]
    \scriptsize
    \centering
    \begin{tabular}{l c c c c c c c c} 
    \toprule
    \multirow{2}{*}{Model} & \multicolumn{4}{c}{MSR-VTT Retrieval} & \multicolumn{4}{c}{DiDeMo Retrieval}  
    \\
    \cmidrule(lr){2-5} \cmidrule(lr){6-9}
    & R@1 $\uparrow$ & R@5 $\uparrow$ & R@10 $\uparrow$ & Mean $\uparrow$ & R@1 $\uparrow$ & R@5 $\uparrow$ & R@10 $\uparrow$ & Mean $\uparrow$ \\
    \cmidrule{1-9}
    CLIP-MeanPool & 43.4 & 70.9 & 81.1 & 65.1 & 40.6 & 67.5 & 77.2 & 61.8 \\
    \cmidrule{1-9}
    CLIP-ViP & 46.5 & 72.1 & 82.5 & 67.0 & 40.6 & 70.4 & 79.3 & 63.4 \\
    $~~~+\mathcal{L}_{V \leftrightarrow S}$ & 47.7 & 72.1 & 82.4 & 67.4 & 44.6 & 73.9 & 81.9 & 66.8 \\ 
    $~~~+\mathcal{L}_{V \leftrightarrow S} + \mathcal{L}_{F \leftrightarrow C}$ & 49.3 & 74.8 & 83.8 & 69.3 & 48.4 & 74.5 & 84.4 & 69.1 \\ 
    $~~~+\mathcal{L}_{V \leftrightarrow S} + \mathcal{L}_{V \leftrightarrow C}$ & 49.6 & 74.2 & 84.0 & 69.3 & 48.5 & 76.6 & 83.6 & 69.5 \\ 
    $~~~+\mathcal{L}_{V \leftrightarrow S} + \mathcal{L}_{V \leftrightarrow C} + \mathcal{L}_{F \leftrightarrow C}$ & 49.6 & 74.5 & 83.8 & 69.3 & 48.5 & 76.8 & 84.1 & 69.8 \\
    $~~~+\mathcal{L}_{V \leftrightarrow S,C} + \mathcal{L}_{F \leftrightarrow C}$ & 49.6 & 74.5 & 84.8 & \bf69.6 & 48.2 & 76.7 & 84.4 & \bf69.8 \\
    
    \bottomrule
    \end{tabular}
    \caption{Ablation study of different losses. We report text-to-video results of models finetuned on MSR-VTT and DiDeMo. Mean $\uparrow$ means the average of Recall@1, 5 and 10. All results are based on CLIP-ViT-B/32. All post-pretraining steps are equivalent to one epoch on HD-VILA-100M.}
    \label{tab:loss}
\end{table*}
}

\newcommand{\tababldata}{
\begin{table*}[t]
    \scriptsize
    \centering
    \begin{tabular}{l c c c c c c c c} 
    \toprule
    \multirow{2}{*}{Post-pretrain Data} & \multicolumn{4}{c}{MSR-VTT Retrieval} & \multicolumn{4}{c}{DiDeMo Retrieval}  
    \\
    \cmidrule(lr){2-5} \cmidrule(lr){6-9}
    & R@1 $\uparrow$ & R@5 $\uparrow$ & R@10 $\uparrow$ & Mean $\uparrow$ & R@1 $\uparrow$ & R@5 $\uparrow$ & R@10 $\uparrow$ & Mean $\uparrow$ \\
    \cmidrule{1-9}
    w/o Post-pretrain & 46.5 & 72.1 & 82.5 & 67.0 & 40.6 & 70.4 & 79.3 & 63.4 \\
    \cmidrule{1-9}
    HD-VILA$_{sub}$ & 47.7 & 72.1 & 82.4 & 67.4 & 44.6 & 73.9 & 81.9 & 66.8 \\ 
    HD-VILA$_{cap}$ & 45.9 & 73.0 & 81.8 & 66.9 & 44.9 & 74.4 & 82.3 & 67.2 \\ 
    HD-VILA$_{sub+cap}$ & \bf49.6 & \bf74.5 & \bf84.8 & \bf69.6 & \bf48.2 & \bf76.7 & \bf84.4 & \bf69.8 \\
    \cmidrule{1-9}
    ImageCaption & 45.6 & 70.7 & 81.1 & 65.8 & 43.7 & 69.5 & 77.9 & 63.7 \\ 
    HD-VILA$_{sub}$ + ImageCaption & 49.1 & 73.1 & 83.5 & 68.6 & 47.0 & 75.3 & 84.1 & 68.8 \\ 
    \bottomrule
    \end{tabular}
    \caption{Ablation study of post-pretrain data. We report text-to-video results of models finetuned on MSR-VTT and DiDeMo. Mean $\uparrow$ indicates an average of Recall@1, 5 and 10. For all results, the model is designed with 4 video proxy tokens and pre-trained on CLIP-ViT-B/32. All post-pretraining steps are equivalent to one epoch on HD-VILA-100M.}
    \label{tab:data}
\end{table*}
}

\newcommand{\figframework}{
\begin{figure*}
    \centering
    \includegraphics[width=\linewidth]{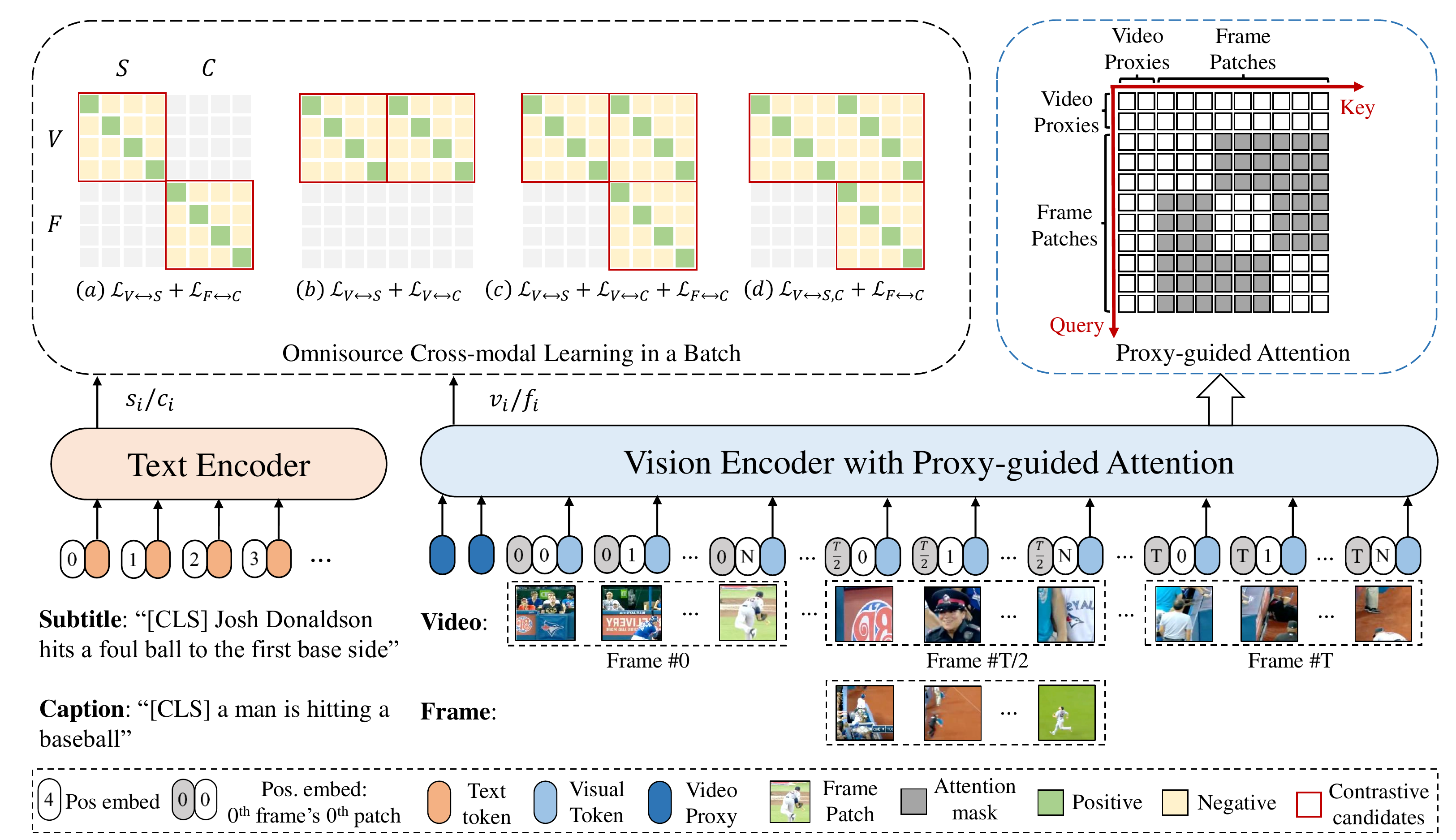}  
    \caption{The framework of CLIP-ViP with a text encoder and a vision encoder. Taken features $V$, $F$, $S$, $C$ of videos, frames, subtitles, captions as input, a series of Omnisource cross-modal learning variants are studied to explore better representation learning losses: (a) $\mathcal{L}_{V \leftrightarrow S} + \mathcal{L}_{F \leftrightarrow C}$; (b) $\mathcal{L}_{V \leftrightarrow S} + \mathcal{L}_{V \leftrightarrow C}$; (c) $\mathcal{L}_{V \leftrightarrow S} + \mathcal{L}_{V \leftrightarrow C} + \mathcal{L}_{F \leftrightarrow C}$; (d) $\mathcal{L}_{V \leftrightarrow S,C} + \mathcal{L}_{F \leftrightarrow C}$. In the vision encoder, Video proxy tokens and the 
    ViP-guided attention mechanism is proposed to transfer CLIP into the video domain. [Best viewed in color]}
    \label{fig:framework}
    \vspace{-5pt}
\end{figure*}
}

\newcommand{\tabmsrvttactnet}{
\begin{table}[t]
    \scriptsize
    \centering
    \begin{tabular}{l c c c c c c c c} 
    \toprule
    \multirow{2}{*}{Method} & \multicolumn{4}{c}{MSR-VTT Retrieval} & \multicolumn{4}{c}{ActivityNet Captions Retrieval}  
    \\
    \cmidrule(lr){2-5} \cmidrule(lr){6-9}
    & R@1 $\uparrow$ & R@5 $\uparrow$ & R@10 $\uparrow$ & Mean $\uparrow$ & R@1 $\uparrow$ & R@5 $\uparrow$ & R@10 $\uparrow$ & Mean $\uparrow$ \\
    \cmidrule{1-9}
    ClipBERT~\citep{lei2021less} & 22.0 & 46.8& 59.9& 6.0  & 21.3 & 49.0 & 63.5 & 6.0\\
    VLM~\citep{xu2021vlm} & 28.1 & 55.5 & 67.4 &4.0 & - & - & - & - \\
    MMT~\citep{gabeur2020multi}  & 26.6 & 57.1 &69.6 & 4.0 & 28.7 & 61.4 & - & 3.3 \\
    Support Set~\citep{patrick2021supportset}  & 30.1 & 58.5& 69.3& 3.0 & 29.2 & 61.6 & - & 3.0 \\
    Frozen~\citep{bain2021frozen}  & 31.0 & 59.5& 70.5& 3.0 & 28.8 & 60.9 & - & 3.0\\
    VideoCLIP~\citep{xu2021videoclip}  & 30.9 & 55.4& 66.8& - & - & - & - & -\\
    HD-VILA~\citep{xue2022advancing}  & 35.6 & 65.3 & 78.0 & 3.0 & 28.5 & 57.4 & - & 4.0 \\
    Florence~\citep{yuan2021florence} & 37.6	& 63.8 & 72.6 & - & - & - & - & -\\
    All-in-One~\citep{wang2022all} & 37.9 & 68.1 & 77.1 & -	& 22.4 & 53.7 & 67.7 & 5.0\\
    BridgeFormer~\citep{ge2022bridging}& 37.6	& 64.8 & 75.1 & 3.0 & - & - & - & -\\
    \cmidrule{1-9}
    \textit{CLIP-ViT-B/32:} & \multicolumn{4}{c}{~} \\
    CLIP4Clip \citep{luo2021clip4clip} & 44.5 &	71.4 &	81.6 &	2.0 & 40.5 & 72.4 &	- &	2.0\\
    CenterCLIP \citep{zhao2022centerclip} & 44.2  & 71.6 & 82.1 & 2.0 & 43.9 & 74.6 & 85.8 & 2.0\\
    XPool \citep{gorti2022x} & 46.9 & 72.8 & 82.2 & 2.0 & - & - & - & -\\
    CLIP2Video \citep{fang2021clip2video}& 45.6 & 72.6 & 81.7	& 2.0 & - & - & - & -\\
    CLIP2Video\textdagger \citep{bogolin2022cross}& 47.2 & 73.0 & 83.0 & 2.0 & - & - & - & -\\
    CLIP2TV  \citep{gao2021clip2tv} & 46.1 & 72.5 & 82.9 & 2.0 & 44.1 & 75.2 & - & 2.0\\
    DRL \citep{wang2022disentangled} & 47.4 & 74.6 & 83.8 & 2.0 & 44.2 & 74.5 & 86.1 & 2.0\\
    CAMoE* \citep{cheng2021improving} & 47.3 & 74.2 & 84.5 & 2.0 & 51.0	& 77.7	& -	& -\\
    Ours & 50.1 & 74.8 & 84.6 & 1.0 & 51.1 & 78.4 & 88.3 & 1.0\\
    Ours* & \bf55.9 & \bf77.0 & \bf86.8 & 1.0 & \bf59.1 & \bf83.9 & \bf91.3 & 1.0\\
    \cmidrule{1-9}
    \textit{CLIP-ViT-B/16:} & \multicolumn{4}{c}{~} \\
    CenterCLIP \citep{zhao2022centerclip} & 48.4 & 73.8 & 82.0 & 2.0 & 46.2 & 77.0 & 87.6 & 2.0\\
    CLIP2TV \citep{gao2021clip2tv} &  49.3 & 74.7 & 83.6 & 2.0 & - & - & - & -\\
    DRL \citep{wang2022disentangled} & 50.2 & 76.5 & 84.7 & 1.0 & 46.2 & 77.3 & 88.2 & 2.0 \\
    DRL\textdagger \citep{wang2022disentangled} & 53.3 & 80.3 & 87.6 & 1.0 & - & - & - & -\\
    Ours & 54.2 & 77.2 & 84.8 & 1.0 & 53.4 & 81.4 & 90.0 & 1.0 \\
    Ours* & \bf57.7 & \bf80.5 & \bf88.2 & 1.0 & \bf61.4 & \bf85.7 & \bf92.6 & 1.0 \\
    \bottomrule
    \end{tabular}
    \caption{Comparison with SOTA models in MSR-VTT~\citep{xu2016msr} and ActivityNet~\citep{Krishna2017actnetcaption} text-to-video retrieval tasks. * and \textdagger ~respectively denotes that the method uses DSL \citep{cheng2021improving} and QB-Norm \citep{bogolin2022cross} as post-processing operations.}
    \label{tab:retrieval-msrvttactnet}
\end{table}
}

\newcommand{\tabactnet}{
\begin{table}[t]
    \scriptsize
    \centering
    \begin{tabular}{l c c c c} 
    \toprule
    Method  & R@1 $\uparrow$ & R@5 $\uparrow$ & R@10 $\uparrow$ & MdR $\downarrow$ 
    \\
    \cmidrule{1-5}
    ClipBERT~\citep{lei2021less} & 21.3 & 49.0 & 63.5 & 6.0  \\
    MMT~\citep{gabeur2020multi}  & 28.7 & 61.4 & - & 3.3 \\
    Support Set~\citep{patrick2021supportset}  & 29.2 & 61.6 & - & 3.0 \\
    Frozen~\citep{bain2021frozen}  & 28.8 & 60.9 & - & 3.0 \\
    HD-VILA~\citep{xue2022advancing}  & 28.5 & 57.4 & - & 4.0 \\
    All-in-One~\citep{wang2022all} & 22.4 & 53.7 & 67.7 & 5.0	\\
    \cmidrule{1-5}
    \textit{CLIP-ViT-B/32} & \multicolumn{4}{c}{~} \\
    CLIP4Clip \citep{luo2021clip4clip} & 40.5 &	72.4 &	- &	2.0 \\
    CenterCLIP \citep{zhao2022centerclip} & 43.9 & 74.6 & 85.8 & 2.0 \\
    CLIP2TV  \citep{gao2021clip2tv} & 44.1 & 75.2 & - & 2.0 \\
    DRL \citep{wang2022disentangled} & 44.2 & 74.5 & 86.1 & 2.0 \\
    CAMoE* \citep{cheng2021improving} & 51.0	& 77.7	& -	& - \\
    Ours & 51.1 & 78.4 & 88.3 & 1.0 \\
    Ours* & \bf59.1 & \bf83.9 & \bf91.3 & 1.0 \\
    \cmidrule{1-5}
    \textit{CLIP-ViT-B/16} & \multicolumn{4}{c}{~} \\
    CenterCLIP \citep{zhao2022centerclip} & 46.2 & 77.0 & 87.6 & 2.0 \\
    DRL \citep{wang2022disentangled} & 46.2 & 77.3 & 88.2 & 2.0 \\
    Ours & 53.4 & 81.4 & 90.0 & 1.0 \\
    Ours* & \bf61.4 & \bf85.7 & \bf92.6 & 1.0 \\
    \bottomrule
    \end{tabular}
    \caption{Comparison with SOTA models in ActivityNet~\citep{Krishna2017actnetcaption}. * denotes that the method uses post-processing operations DSL \citep{cheng2021improving}.}
    \label{tab:retrieval-actnet}
\end{table}
}

\newcommand{\tabmsrvtt}{
\begin{table}[t]
    \scriptsize
    \centering
    \begin{tabular}{l c c c c} 
    \toprule
    Method  & R@1 $\uparrow$ & R@5 $\uparrow$ & R@10 $\uparrow$ & MdR $\downarrow$ 
    \\
    \cmidrule{1-5}
    ClipBERT~\citep{lei2021less} & 22.0 & 46.8& 59.9& 6.0  \\
    VLM~\citep{xu2021vlm} & 28.1 & 55.5 & 67.4 &4.0  \\
    MMT~\citep{gabeur2020multi}  & 26.6 & 57.1 &69.6 & 4.0 \\
    Support Set~\citep{patrick2021supportset}  & 30.1 & 58.5& 69.3& 3.0 \\
    Frozen~\citep{bain2021frozen}  & 31.0 & 59.5& 70.5& 3.0 \\
    VideoCLIP~\citep{xu2021videoclip}  & 30.9 & 55.4& 66.8& - \\
    HD-VILA~\citep{xue2022advancing}  & 35.6 & 65.3 & 78.0 & 3.0 \\
    Florence~\citep{yuan2021florence} & 37.6	& 63.8 & 72.6 & - \\
    All-in-One~\citep{wang2022all} & 37.9 & 68.1 & 77.1 & -	\\
    BridgeFormer~\citep{ge2022bridging}& 37.6	& 64.8 & 75.1 & 3.0 \\
    \cmidrule{1-5}
    \textit{CLIP-ViT-B/32} & \multicolumn{4}{c}{~} \\
    CLIP4Clip \citep{luo2021clip4clip} & 44.5 &	71.4 &	81.6 &	2.0 \\
    CenterCLIP \citep{zhao2022centerclip} & 44.2  & 71.6 & 82.1 & 2.0 \\
    XPool \citep{gorti2022x} & 46.9 & 72.8 & 82.2 & 2.0 \\
    CLIP2Video \citep{fang2021clip2video}& 45.6 & 72.6 & 81.7	& 2.0 \\
    CLIP2Video\textdagger \citep{bogolin2022cross}& 47.2 & 73.0 & 83.0 & 2.0 \\
    CLIP2TV  \citep{gao2021clip2tv} & 46.1 & 72.5 & 82.9 & 2.0 \\
    DRL \citep{wang2022disentangled} & 47.4 & 74.6 & 83.8 & 2.0 \\
    CAMoE* \citep{cheng2021improving} & 47.3 & 74.2 & 84.5 & 2.0\\
    Ours & 50.1 & 74.8 & 84.6 & 1.0 \\
    Ours* & \bf55.9 & \bf77.0 & \bf86.8 & 1.0 \\
    \cmidrule{1-5}
    \textit{CLIP-ViT-B/16} & \multicolumn{4}{c}{~} \\
    CenterCLIP \citep{zhao2022centerclip} & 48.4 & 73.8 & 82.0 & 2.0 \\
    CLIP2TV \citep{gao2021clip2tv} &  49.3 & 74.7 & 83.6 & 2.0 \\
    DRL \citep{wang2022disentangled} & 50.2 & 76.5 & 84.7 & 1.0 \\
    DRL\textdagger \citep{wang2022disentangled} & 53.3 & 80.3 & 87.6 & 1.0 \\
    Ours & 54.2 & 77.2 & 84.8 & 1.0 \\
    Ours* & \bf57.7 & \bf80.5 & \bf88.2 & 1.0 \\
    \bottomrule
    \end{tabular}
    \caption{Comparison of text-to-video retrieval in MSR-VTT~\citep{xu2016msr}. * and \textdagger ~respectively denotes that the method uses DSL \citep{cheng2021improving} and QB-Norm \citep{bogolin2022cross} as post-processing operations.}
    \label{tab:retrieval-msrvtt}
\end{table}
}

\newcommand{\tabDiDeMo}{
\begin{table}[t]
    \scriptsize
    \centering
    \begin{tabular}{l c c c c} 
    \toprule
    Method  & R@1 $\uparrow$ & R@5 $\uparrow$ & R@10 $\uparrow$ & MdR $\downarrow$ 
    \\
    \cmidrule{1-5}
    ClipBERT~\citep{lei2021less} & 20.4 & 48.0 & 60.8 & 6.0  \\
    Frozen~\citep{bain2021frozen}  & 31.0 &  59.8 & 72.4  & 3.0 \\
    HD-VILA~\citep{xue2022advancing}  & 28.8 & 57.4 & 69.1 & 4.0 \\
    All-in-One~\citep{wang2022all} & 32.7 & 61.4 & 73.5 & 3.0	\\
    BridgeFormer~\citep{ge2022bridging}& 37.0 & 62.2 & 73.9 & 3.0 \\
    \cmidrule{1-5}
    \textit{CLIP-ViT-B/32} & \multicolumn{4}{c}{~} \\
    CLIP4Clip \citep{luo2021clip4clip} & 43.4 & 70.2 & 80.6 & 2.0 \\
    CLIP2TV  \citep{gao2021clip2tv} & 45.5 & 69.7 & 80.6 & 2.0  \\
    DRL \citep{wang2022disentangled} & 47.9 & 73.8 & 82.7 & 2.0 \\
    CAMoE* \citep{cheng2021improving} & 43.8 & 71.4 & - & -\\
    Ours & 48.6 & 77.1 & 84.4 & 2.0 \\
    Ours* & \bf53.8 & \bf79.6 & \bf86.5 & 1.0 \\
    \cmidrule{1-5}
    \textit{CLIP-ViT-B/16} & \multicolumn{4}{c}{~} \\
    DRL \citep{wang2022disentangled} & 49.0 & 76.5 & 84.5 & 2.0 \\
    Ours & 50.5 & 78.4 & 87.1 & 1.0 \\
    Ours* & \bf55.3 & \bf82.0 & \bf89.3 & 1.0 \\
    \bottomrule
    \end{tabular}
    \caption{Comparison of text-to-video retrieval in DiDeMo~\citep{anne2017localizing}. * denotes that the method uses post-processing operations DSL \citep{cheng2021improving}.}
    \label{tab:retrieval-DiDeMo}
\end{table}
}

\newcommand{\tabDiDeMolsmdc}{
\begin{table}[t]
    \scriptsize
    \centering
    \begin{tabular}{l c c c c c c c c} 
    \toprule
    \multirow{2}{*}{Method} & \multicolumn{4}{c}{DiDeMo Retrieval} & \multicolumn{4}{c}{LSMDC Retrieval}  
    \\
    \cmidrule(lr){2-5} \cmidrule(lr){6-9}
    & R@1 $\uparrow$ & R@5 $\uparrow$ & R@10 $\uparrow$ & Mean $\uparrow$ & R@1 $\uparrow$ & R@5 $\uparrow$ & R@10 $\uparrow$ & Mean $\uparrow$ \\
    \cmidrule{1-9}
    MMT~\citep{gabeur2020multi}  & - & - & - & - & 12.9 & 29.9 &40.1 & 19.3 \\
    ClipBERT~\citep{lei2021less} & 20.4 & 48.0 & 60.8 & 6.0 & - & - & - & - \\
    Frozen~\citep{bain2021frozen}  & 31.0 &  59.8 & 72.4  & 3.0 & 15.0 & 30.8 & 40.3 & 20.0  \\
    HD-VILA~\citep{xue2022advancing}  & 28.8 & 57.4 & 69.1 & 4.0 & 17.4 & 34.1 & 44.1 & 15.0  \\
    All-in-One~\citep{wang2022all} & 32.7 & 61.4 & 73.5 & 3.0 & - & - & - & -	\\
    BridgeFormer~\citep{ge2022bridging}& 37.0 & 62.2 & 73.9 & 3.0 & 17.9 & 35.4 & 44.5 & 15.0\\
    \cmidrule{1-9}
    \textit{CLIP-ViT-B/32:} & \multicolumn{4}{c}{~} \\
    CLIP4Clip \citep{luo2021clip4clip} & 43.4 & 70.2 & 80.6 & 2.0 & 21.6 & 41.8 & 49.8 & 11.0 \\
    CenterCLIP \citep{zhao2022centerclip} & - & - & - & - & 21.7 & 39.8 & 49.8 & 11.0\\
    XPool \citep{gorti2022x} & - & - & - & - & 22.7 & 42.6 & 51.2 & 10.0 \\
    CLIP2TV  \citep{gao2021clip2tv} & 45.5 & 69.7 & 80.6 & 2.0 & - & - & - & - \\
    DRL \citep{wang2022disentangled} & 47.9 & 73.8 & 82.7 & 2.0 & 24.9 & 45.7 & \bf55.3 & 7.0 \\
    CAMoE* \citep{cheng2021improving} & 43.8 & 71.4 & - & - & 25.9 & 46.1 & 53.7 & -\\
    Ours & 48.6 & 77.1 & 84.4 & 2.0 & 25.6 & 45.3 & 54.4 & 8.0 \\
    Ours* & \bf53.8 & \bf79.6 & \bf86.5 & 1.0 & \bf26.0 & \bf46.4 & 54.9 & 8.0 \\
    \cmidrule{1-9}
    \textit{CLIP-ViT-B/16:} & \multicolumn{4}{c}{~} \\
    CLIP4Clip by \citep{zhao2022centerclip} & - & - & - & -  & 24.1 & 45.0 & 55.1 & 8.0 \\
    CenterCLIP \citep{zhao2022centerclip} & - & - & - & - & 24.2 & 46.2 & 55.9 & 8.0 \\
    DRL \citep{wang2022disentangled} & 49.0 & 76.5 & 84.5 & 2.0 & 26.5 & 47.6 & 56.8 & 7.0 \\
    Ours & 50.5 & 78.4 & 87.1 & 1.0 & 29.4 & 50.6 & 59.0 & 5.0 \\
    Ours* & \bf55.3 & \bf82.0 & \bf89.3 & 1.0 & \bf30.7 & \bf51.4 & \bf60.6 & 5.0 \\
    \bottomrule
    \end{tabular}
    \caption{Comparison with SOTA models in DiDeMo~\citep{anne2017localizing} and LSMDC~\citep{Rohrbach2016MovieD} text-to-video retrieval tasks. * denotes using post-processing DSL \citep{cheng2021improving}.}
    \label{tab:retrieval-DiDeMolsmdc}
\end{table}
}

\newcommand{\tablsmdc}{
\begin{table}[t]
    \scriptsize
    \centering
    \begin{tabular}{l c c c c} 
    \toprule
    Method  & R@1 $\uparrow$ & R@5 $\uparrow$ & R@10 $\uparrow$ & MdR $\downarrow$ 
    \\
    \cmidrule{1-5}
    MMT~\citep{gabeur2020multi}  & 12.9 & 29.9 &40.1 & 19.3 \\
    Frozen~\citep{bain2021frozen}  & 15.0 & 30.8 & 40.3 & 20.0 \\
    HD-VILA~\citep{xue2022advancing}  & 17.4 & 34.1 & 44.1 & 15.0 \\
    BridgeFormer~\citep{ge2022bridging}& 17.9 & 35.4 & 44.5 & 15.0 \\
    \cmidrule{1-5}
    \textit{CLIP-ViT-B/32} & \multicolumn{4}{c}{~} \\
    CLIP4Clip \citep{luo2021clip4clip} & 21.6 & 41.8 & 49.8 & 11.0 \\
    CenterCLIP \citep{zhao2022centerclip} & 21.7 & 39.8 & 49.8 & 11.0\\
    XPool \citep{gorti2022x} & 22.7 & 42.6 & 51.2 & 10.0 \\
    DRL \citep{wang2022disentangled} & 24.9 & 45.7 & \bf55.3 & 7.0 \\
    CAMoE* \citep{cheng2021improving} & 25.9 & 46.1 & 53.7 & - \\
    Ours & 25.6 & 45.3 & 54.4 & 8.0 \\
    Ours* & \bf26.0 & \bf46.4 & 54.9 & 8.0 \\
    \cmidrule{1-5}
    \textit{CLIP-ViT-B/16} & \multicolumn{4}{c}{~} \\
    CLIP4Clip by \citep{zhao2022centerclip} & 24.1 & 45.0 & 55.1 & 8.0 \\
    CenterCLIP \citep{zhao2022centerclip} & 24.2 & 46.2 & 55.9 & 8.0 \\
    DRL \citep{wang2022disentangled} & 26.5 & 47.6 & 56.8 & 7.0 \\
    Ours & 29.4 & 50.6 & 59.0 & 5.0 \\
    Ours* & \bf30.7 & \bf51.4 & \bf60.6 & 5.0 \\
    \bottomrule
    \end{tabular}
    \caption{Comparison of text-to-video retrieval in LSMDC~\citep{Rohrbach2016MovieD}. * denotes that the method uses post-processing operations DSL \citep{cheng2021improving}.}
    \label{tab:retrieval-lsmdc}
\end{table}
}

\newcommand{\tabmsvd}{
\begin{table}[t]
    \scriptsize
    \centering
    \begin{tabular}{l c c c c} 
    \toprule
    Method  & R@1 $\uparrow$ & R@5 $\uparrow$ & R@10 $\uparrow$ & MdR $\downarrow$ 
    \\
    \cmidrule{1-5}
    Support Set~\citep{patrick2021supportset}  & 28.4 & 60.0& 72.9 & 4.0 \\
    Frozen~\citep{bain2021frozen}  & 33.7 & 64.7& 76.3 & 3.0 \\
    \cmidrule{1-5}
    \textit{CLIP-ViT-B/32} & \multicolumn{4}{c}{~} \\
    CLIP4Clip \citep{luo2021clip4clip} & 46.2 &	76.1 &	84.6 &	2.0 \\
    CenterCLIP \citep{zhao2022centerclip} & 47.6  & 77.5 & 86.0 & 2.0 \\
    XPool \citep{gorti2022x} & 47.2 & 77.4 & 86.0 & 2.0 \\
    CLIP2Video \citep{fang2021clip2video}& 47.0 & 76.8 & 85.9	& 2.0 \\
    CLIP2Video* \citep{bogolin2022cross}& 48.0 & 77.9 & 86.2 & 2.0 \\
    DRL \citep{wang2022disentangled} & 48.3 & 79.1 & 87.3 & 2.0 \\
    CAMoE* \citep{cheng2021improving} & 49.8 & 79.2 & 87.0 & - \\
    Ours & - & - & - & - \\
    Ours* & - & - & - & - \\
    \cmidrule{1-5}
    \textit{CLIP-ViT-B/16} & \multicolumn{4}{c}{~} \\
    CLIP4Clip by \citep{zhao2022centerclip} & 49.6 & 79.5 & 88.0 & 2.0 \\
    CenterCLIP \citep{zhao2022centerclip} & 50.6 & 80.3 & 88.4 & 1.0 \\
    DRL \citep{wang2022disentangled} &  50.0 & 81.5 & 89.5 & 2.0 \\
    Ours & - & - & - & - \\
    Ours* & - & - & - & - \\
    \bottomrule
    \end{tabular}
    \caption{Comparison of text-to-video retrieval in MSVD~\citep{chen2011collecting}. * denotes that the method uses post-processing operations DSL \citep{cheng2021improving}.}
    \label{tab:retrieval-msvd}
\end{table}
}

\begin{abstract}
% The pre-trained image-text models like CLIP, have demonstrated the big power of vision-language representation learning from web-collected large scale image-text pairs. In light of well-learned visual concepts, some existing works transfer CLIP  to video domain and achieve good results. However, post-pretraining CLIP on video data is still under explored. In this paper, we investigate two questions: 1) what are the factors hindering post-pretraining CLIP to further improve the performance on video-language tasks? and 2) how to mitigate the impact of these factors? Through series of comparison experiments and analyses, we find that the data scale and domain gap between language sources have great impacts. Motivated by these, we propose a multi-source post-pretraining method equipped with a video-clue mechanism. Extensive results show that our approach improves the performance of CLIP on video-text retrieval by a large margin. Our model also achieves SOTA results on a variety of datasets, including MSR-VTT, DiDeMo, LSMDC, and ActivityNet. We release our code at \url{https://github.com/XXX}.
Pre-trained image-text models, like CLIP, have demonstrated the strong power of vision-language representation learned from a large scale of web-collected image-text data. In light of the well-learned visual features, there are works that transfer image representation to the video domain and achieve good results. 
% However, how to utilize the image-text pre-trained models (e.g., CLIP) for video-text pre-training (post-pretraining) is still under-explored. 
However, adapting image-text pre-trained models to video-text pre-training (i.e., post-pretraining) has not demonstrated a significant advantage yet.
In this paper, we tackle this challenge by raising and addressing two questions: 1) what are the factors hindering post-pretraining CLIP from improving performance on video-text tasks, and 2) how to mitigate the impact of these factors. Through a series of comparative experiments and analyses, we find that the data scale and domain gap between language sources have large impacts. By these observations, we propose an Omnisource Cross-modal Learning method equipped with a \textbf{Vi}deo \textbf{P}roxy mechanism on the basis of CLIP, namely CLIP-ViP. Extensive results show that our approach improves the performance of CLIP on video-text retrieval by a large margin. Our model achieves state-of-the-art results on a variety of datasets, including MSR-VTT, DiDeMo, LSMDC, and ActivityNet. We release our code and pre-trained CLIP-ViP models at \url{https://github.com/microsoft/XPretrain/tree/main/CLIP-ViP}.

\end{abstract} 
\section{Introduction} \label{introduction}
% In the last few years, vision-language pre-training has achieved great success on cross-modal representation learning from web-crawled data \citep{radford2021learning,jia2021scaling,Li2021AlignBF,wang2021simvlm,zellers2021merlot,zellers2022merlot,bain2021frozen}. Among them, some work pre-train on large scale image alt-text pairs \citep{radford2021learning,jia2021scaling} and start to show unprecedented power on varies downstream tasks \citep{patashnik2021styleclip,zhang2022pointclip,mokady2021clipcap,gu2021open}. In light of well-learned connections between visual concepts and language, some work directly transfer image-text pre-trained model to video-text retrieval without further pre-training on video data \citep{luo2021clip4clip,fang2021clip2video,gorti2022x,zhao2022centerclip}, while outperforming models pre-trained on data containing videos \citep{xu2021videoclip,bain2021frozen,xue2022advancing}.
In the past few years, vision-language pre-training has achieved great success on cross-modal representation learning from a large scale of web-crawled data \citep{radford2021learning,jia2021scaling,Li2021AlignBF,wang2021simvlm,zellers2021merlot,zellers2022merlot,bain2021frozen}. Among them, image-text pre-trained models \citep{radford2021learning,jia2021scaling} have shown powerful capability for various downstream tasks, including visual understanding \citep{gu2021open,wang2021actionclip,rao2022denseclip} , image-text generation \citep{patashnik2021styleclip,mokady2021clipcap} and so on \citep{guzhov2022audioclip,zhang2022pointclip}. In light of the well-learned and enriched visual representation, some works directly adapt image-text pre-trained models to video-text downstream tasks without further pre-training on video data \citep{luo2021clip4clip,fang2021clip2video,gorti2022x,zhao2022centerclip}, while still outperforming models pre-trained on video data \citep{xu2021videoclip,bain2021frozen}.

% Further pre-training an existing image-language model on video data (video post-pretraining) not only helps to reduce the total training cost, but also makes good use of the knowledge already learned from images. However, video post-pretraining has not yet demonstrated significant advantages thus is still under explored. A preliminary study is first conducted by CLIP4Clip \citep{luo2021clip4clip}. They optimize the CLIP in MeanPool type on a subset of Howto100M \citep{miech2019howto100m}. Evaluating on retrieval, the performance improves a bit on zero-shot, but minimally  on fine-tuning. In this paper, we aim at improving image-language models' performance on video-language tasks. We adopt CLIP and mainly focus on text-to-video retrieval task on various datasets. 
Utilizing an existing powerful image-text pre-trained model for further video-text pre-training (i.e., post-pretraining) is able to reduce the required training cost by making good use of the knowledge learned from images. However, adapting image-text pre-trained models to video-text data for post-pretraining has not demonstrated a significant advantage yet, and thus is still under-explored. 
A preliminary study is conducted by CLIP4Clip \citep{luo2021clip4clip} which adopts MeanPooling by averaging multiple frame features based on the CLIP model on a subset of Howto100M \citep{miech2019howto100m}. While the improvement over directly using the image-text pre-trained model is marginal for either zero-shot or fine-tuning settings. 
In this paper, we aim to explore how to effectively adapt the image-text pre-trained model (e.g., CLIP) to video-language representation learning for video-text tasks (e.g., text-to-video retrieval). 

% To unleash the power of video data for post-pretraining CLIP, we conduct a preliminary analysis to figure out what factors hinder post-pretraining. First, we study post pre-training a CLIP model in MeanPool type on existing video-language datasets, including WebVid-2.5M \citep{bain2021frozen} and HD-VILA-100M \citep{xue2022advancing}. The results show that the scale of the data is important. Small-scale data makes the model tend to over-fit the new data thus forget the implicit knowledge learned from the image-text pairs, which in turn reduces the model's performance. Second, we study the domain gap between pre-training data and the downstream task data. We achieve this by calculating the Normalized Mutual Information (NMI) on clusters of text features. The results show that there is a large domain gap between current video-subtitle data and downstream task data.
To unleash the power of video data to adapt image-text pre-trained models for post-pretraining, we conduct several preliminary experiments to figure out the challenges that hinder post-pretraining. First, we explore post-pretraining an image-text pre-trained model (i.e., CLIP) with MeanPooling on video-text datasets with different scales, including WebVid-2.5M \citep{bain2021frozen} and HD-VILA-100M \citep{xue2022advancing}. The result shows that the scale of data is critical for video-text post-pretraining. Data on a small scale makes the model easy to over-fit the new data while the knowledge learned from image-text is suppressed and the performance is reduced. Second, we investigate the language domain gap between pre-training data and downstream data. By calculating the Normalized Mutual Information (NMI) on clusters of text features, we find that there is a large domain gap between subtitles that are used in large-scale video-text pre-training data and descriptive texts in downstream tasks.

To mitigate the impact of the above factors, we propose CLIP-ViP to adapt the pre-trained image-text model CLIP for video-text pre-training. 
First, we introduce auxiliary captions that have a smaller language domain gap with downstream data into existing large-scale video-text data. 
Instead of using a video captioning model which may cause data leakage by training on the same dataset with video-text downstream tasks, and considering a better visual captioning capability, we adopt an image captioning model to generate an auxiliary caption of middle frame in each video.
In order to adapt a Transformer-based vision encoder to process both images and videos with minimal modification, we then propose video proxy tokens and design a proxy-guided video attention  mechanism for the Vision Transformer (ViT). Specifically, during attention computation in each block, video proxy tokens can interact with all tokens, while patch tokens only interact with video proxy tokens and patch tokens within the same frame. Our vision encoder only increases negligible parameters and calculations compared to the vanilla Vision Transformer while increasing the generality and extendability.  
To facilitate cross-modal representation learning from both caption-frame and video-subtitle data types at the same time, we propose an Omnisource Cross-modal Learning (OCL) method for pre-training and study a series of variants to find the best fusion strategy.

Our experimental results show that our approach improves the performance of CLIP on text-to-video retrieval tasks by a large margin. We also conduct ablation studies to verify the effectiveness of each part in our approach.
% Our contributions are summarized as follows: (1) We are one of the first to conduct preliminary analysis and reveal two factors that hinder video post-pretraining on pre-trained image-text models: data scale and language domain gap; (2) We then propose CLIP-ViP, a video-language pre-training model equipped with a Video Proxy mechanism and an Omnisource Cross-modal learning method to learn from both image-text and video-text pairs; and 
% (3) We conduct extensive experiments to verify the effectiveness of our method, which outperforms the state-of-the-art results by a large margin on a variety of datasets.
Our contributions are summarized as follows: (1) We are one of the first to explore factors that hinder video post-pretraining on pre-trained image-text models; (2) We propose CLIP-ViP that can effectively leverage image-text pre-trained model for post-pretraining;
(3) We conduct extensive experiments to verify the effectiveness of our method. Our model outperforms the state-of-the-art results by a large margin on four widely-used benchmarks.

\vspace{-5pt}
\section{Related Work}
\vspace{-5pt}
\paragraph{Vision-Language Pre-Training}
End-to-end models \citep{lei2021less,xue2022advancing,zellers2021merlot,fu2021violet,huang2020pixel,huang2021seeing,xue2021probing,Li2021AlignBF,kim2021vilt,huang2021unifying,sun2022long} for vision-language pre-training are replacing the traditional approach using pre-extracted visual features by off-the-shelf models \citep{sun2019videobert,xu2021videoclip,zhu2020actbert,li2020oscar,li2020unicoder,chen2020uniter}. Training end-to-end models on large-scale web-collected data also gradually demonstrates the big advantages \citep{radford2021learning,jia2021scaling,xue2022advancing,zellers2021merlot,zellers2022merlot}. Unlike images that have alt-texts, large-scale video datasets suitable for pre-training usually use subtitles as text sources \citep{miech2019howto100m,xue2022advancing}. Subtitles are much noisier than alt-texts, according to \citep{miech2019howto100m}, typical examples of incoherence include the content producer asking viewers to subscribe to their channel, talking about something unrelated to the video, or describing something before or after it happens. Bain \textit{et al.} collect a video dataset WebVid \citep{bain2021frozen} with textual description annotations. Their texts are well aligned with the video and avoid suffering from ASR errors. However, the vast majority of WebVid videos are sourced from a stock footage website, so scaling up is under limitation. The video-subtitle data is more easily accessible on the web and thus suitable for scaling up. In this paper, we investigate the unfavorable factors of video-subtitle data and explore how to mitigate the impact of these factors.

\paragraph{Pre-trained Models for Video-Text Retrieval}
The great success of the CLIP has demonstrated its unprecedented power on varies downstream tasks, including vision understanding \citep{gu2021open,wang2021actionclip,rao2022denseclip}, image-text generation \citep{patashnik2021styleclip,mokady2021clipcap} and so on \citep{guzhov2022audioclip,zhang2022pointclip}. By contrastive learning on large-scale image-text pairs, CLIP learns enriched visual concepts for images. Recently, some works directly transfer CLIP to video-text retrieval without further pretraining on video data (post-pretraining) \citep{luo2021clip4clip,fang2021clip2video,gorti2022x,zhao2022centerclip,wang2022disentangled}. Their work takes the performance of video-text retrieval to a new level, outperforming existing models pre-trained on video data \citep{xu2021videoclip,bain2021frozen,xue2022advancing,ge2022bridging,wang2022all}. They transfer CLIP from views of feature aggregation \citep{luo2021clip4clip,zhao2022centerclip,fang2021clip2video,gorti2022x} or representation alignment \citep{fang2021clip2video,gorti2022x,wang2022disentangled}. In parallel with these works, we study post-pretraining with video data on top of CLIP in an effective way and our model can be combined with other approaches effectively. 
\vspace{-5pt}
\section{Preliminary Analysis}\label{sec:analysis}
\vspace{-5pt}
\figmerge

%  In this section, we first study the impact of data scale on post-pretraining, then explore the unfavorable factors of large-scale data from the language domain's perspective. 
In this section, we first study the impact of the data scale for adapting image-text pre-training to video-text post-pretraining,
% Then we explore the unfavorable factors of large-scale data from the vision-aware language domain's perspective. 
and then investigate how the language domain gap affects the adaption.
\vspace{-5pt}
\subsection{Post-pretraining with Different Data Scales}
\vspace{-5pt}
% We directly post-pretrain a CLIP-ViT-B/32 model in meanPool setting on typical datasets for video-language pre-training: WebVid-2.5M \citep{}, HD-VILA-100M \citep{}, and a subset of HD-VILA-100M containing random 10\% data. Notice that we use the low-resolution version of HD-VILA-100M for all experiments in this paper. We run the same number of steps on all datasets, equivalent to 40 epochs on WebVid-2.5M. We uniformly sample 12 frames for each video and apply the same hyper-parameters as described in Section \ref{sec:exp} for all datasets.
To study the effectiveness of different data scales, we use the CLIP-ViT-B/32 model \citep{radford2021learning} as the base image-text pre-trained model and adopt MeanPooling for video adaption like CLIP4Clip \citep{luo2021clip4clip} by averaging multiple frame features as video feature. Two open-domain video-text datasets are used: WebVid-2.5M \citep{bain2021frozen} with 2.5 million pairs and HD-VILA-100M \citep{xue2022advancing} with 100M pairs. We also adopt a subset of HD-VILA-100M containing random 10\% data (namely HD-VILA-10M) as a middle setting. We run the same number of steps on all settings, equivalent to one epoch on HD-VILA-100M. We uniformly sample 12 frames from each video and apply the same hyper-parameters as described in Section \ref{sec:exp} for all settings.

% During post-pretraining, we evaluate by finetuning on MSR-VTT text-to-video retrieval. From the results in Figure \ref{fig:precurve}, We observe a overfitting phenomenon that continuous post-pretraining leads to a decrease in performance. And the this phenomenon is more significant on smaller data. As CLIP is pre-trained on 400 million image-text pairs, further training on small data makes the model tend to overfit the new data thus forget the implicit knowledge learned from the image-text pairs. As a consequence, the performance will drop, even worse than using CLIP directly. This motivate us to adopt HD-VILA-100M due to its large scale and easy scalability.
During post-pretraining, we evaluate the pre-trained models by fine-tuning on MSR-VTT text-to-video retrieval task. Figure \ref{fig:precurve} shows the performance trend. We observe an overfitting phenomenon that continuous post-pretraining leads to a performance drop. And the drop is more significant for smaller data (e.g., WebVid-2.5M and HD-VILA-10M). As CLIP is pre-trained on 400 million image-text pairs, further training on small data makes the model tend to overfit the new data while the implicit knowledge learned from the image-text pairs is degrading. As a consequence, the performance will drop, even worse than using CLIP directly. Thus we adopt HD-VILA-100M due to its large scale and diverse category.
\vspace{-5pt}
\subsection{Language Domain Gap with Downstream data}
\vspace{-5pt}
% Video-subtitle data suffers from noise from video irrelevance, misalignment, ASR error. On the other hand, subtitles are very inconsistent with the language of the caption-style texts used in most downstream tasks. We study this inconsistency by calculating the dissimilarity of the language features. We choose two typical video-text retrieval datasets: MSR-VTT \citep{xu2016msr} and DiDeMo \citep{anne2017localizing}. We choose HD-VILA-100M's subtitles (HD-VILA$_{sub}$), WebVid's manual captions, MS-COCO's image captions \citep{lin2014microsoft}, and Conceptual Caption 12M's web-collected alt-texts \citep{changpinyo2021conceptual} as comparison data. Besides, we add auto-generated captions of HD-VILA-100M (HD-VILA$_{cap}$), which will be described in Section \ref{sec:approach}.
It is intuitive that pre-training on data with the same domain as downstream data can benefit downstream tasks. For most video-text tasks like video-text retrieval, texts are descriptive sentences of videos (i.e., captions). While for HD-VILA-100M, which we will use for pre-training, the texts are auto-transcribed subtitles and they indicate very different relevance to visual information compared to descriptive texts. Meanwhile, auto-transcribed subtitles suffer from irrelevance, misalignment, and ASR errors \citep{miech2019howto100m}. To better explore the language domain gap between pre-training data and downstream data, we measure the inconsistency by calculating the dissimilarity between their language features.
For downstream language data, we choose two typical video-text retrieval datasets: MSR-VTT \citep{xu2016msr} and DiDeMo \citep{anne2017localizing}. For pre-training language, we select four types: video subtitles of HD-VILA-100M (HD-VILA$_{sub}$), video captions of WebVid-2.5M, image captions of MS-COCO \citep{lin2014microsoft}, and web-collected alt-texts of Conceptual Caption 12M \citep{changpinyo2021conceptual}. In addition, we analyze auto-generated captions of HD-VILA-100M (HD-VILA$_{cap}$), which will be introduced in Section \ref{sec:approach}.

% To extract language features, we use a Transformer Encoder initialized from BERT \citep{kenton2019bert} or CLIP \citep{radford2021learning}. To calculate domain gap, we mix language features from two datasets then use K-means to form two clusters. We calculate the Normalized Mutual Information (NMI) between K-means cluster label and ground-truth. A larger NMI value means that the two types of features are more dissimilar. For each data type, we randomly sample 1000 texts to extract features, and all results are the average of 10 experiments. We report the results in Table \ref{tab:analysis_nmi}. We mainly focus on the CLIP-init results as we want to study vision-aware language features. BERT-init results provide references from a language-only aspect. From the last column, we find that the NMI score between HD-VILA$_{sub}$ and downstream data is much larger than others, especially MSR-VTT. This indicates that direct training with subtitles may introduce inconsistencies with downstream tasks.
We use a Transformer Encoder initialized from CLIP \citep{radford2021learning} to extract text features. To quantify the domain gap of languages between pre-training and downstream data, we first mix their text features and then use K-means to get two clusters. Then we calculate the Normalized Mutual Information (NMI) between cluster labels and ground-truth labels of pre-training or downstream. A larger NMI value means that the two types of features are easy to be distinguished, thus there is a larger domain gap. For each comparison, we randomly sample 1000 texts from each type of data for 10 times and adopt the average of 10 results. We report the results in Table \ref{tab:analysis_nmi}. Comparing the values of all pre-training data types, we find that the NMI score between HD-VILA$_{sub}$ and downstream data is much larger than others, especially for MSR-VTT downstream dataset. This indicates that direct training with subtitles may introduce inconsistency with downstream tasks. 

% \figprecurve
% \tabprenmi
\vspace{-5pt}
\section{Approach}\label{sec:approach}
\vspace{-5pt}
\figframework
\vspace{-8pt}
In this section, we will introduce the proposed CLIP-ViP video pre-training framework. To bridge language domain gaps between image and video datasets, we first introduce an in-domain auxiliary data generation method. Then, we propose a novel Video Proxy mechanism to enable the Vision Transformer (ViT) model for both image and video encoding. We further present an Omnisource Cross-modal Learning (OCL) method which can jointly learn cross-modal representation from video-text and image-text pairs. 

% \subsection{Auxiliary Captions}
\vspace{-5pt}
\subsection{In-domain Auxiliary Data Generation}
\vspace{-5pt}
% Motivated by the analysis in Section \ref{sec:analysis}, we introduce auxiliary captions into large-scale video-subtitle data to reduce the domain gap. We adopt image captioning model for two reasons: 1) The training dataset of current video captioning models are included in our downstream tasks, e.g. MSR-VTT.  We avoid data leakage to perform pre-training agnostic to downstream data. 2) The performance of existing captioning models on videos lags far behind that on images. For above reasons, we choose a powerful image captioning model OFA-Caption \citep{wang2022unifying} to generate one caption for the middle frame of each video in HD-VILA-100M. We use the default setting of OFA-Caption model. This approach can be applied to any video data and we will also release auxiliary captions on HD-VILA-100M to facilitate future research.
Motivated by the analysis in Section \ref{sec:analysis}, we introduce auxiliary captions into large-scale video-subtitle data to reduce the language domain gap between pre-training and downstream data. We adopt an image captioning model for two reasons. 1) Most SOTA video captioning models are trained with video-text datasets (e.g., MSR-VTT, ActivityNet) which are also used for downstream tasks. We avoid data leakage to perform pre-training agnostic to downstream data. 2) The performance of existing video captioning models lags far behind that of images. Thus, we choose a powerful image captioning model OFA-Caption \citep{wang2022unifying} to generate one caption for the middle frame of each video in HD-VILA-100M. We use the default setting of the OFA-Caption model. 
As a result, we generate 100M sentences with a max length of 16 words. 
% The gererated sentences cover a vocabulary of XXX and have a overlap of XX\% with subtitles of HD-VILA-100M.
This method can be applied to any video data and we will release the generated captions to facilitate future research.

\vspace{-5pt}
\subsection{Video Proxy Mechanism}
\vspace{-5pt}\label{sec:vip}
Since video is an ordered sequence of frames, it is critical to learn the frame aggregation and temporality when transferring to the video domain. Meanwhile, to keep the high generality and extendability of the Vision Transformer (ViT) backbone, we aim to find a simple but effective way to transfer ViT to enable both image and video encoding with minimal modifications.
Given a video containing $T$ frames: $\{f_1, f_2, ..., f_T\}$, we follow CLIP to divide each frame into N patches: $\{f^1_t, f^2_t, ..., f^N_t ~|~ t \in [1, T]\}$. Then we add spatio-temporal positional embedding to each  flattened 2D patches:
\begin{equation}
\begin{aligned}
g(f^n_t) = Linear(f^n_t) + Pos_{s}(n) + Pos_{t}(t),
\end{aligned}
\end{equation}
where $Linear(*)$ is a linear layer, $Pos_{s}(n)$ and $Pos_{t}(t)$ is the learnable spatial and temporal positional embedding, respectively. The whole video can be divided into $T\times N$ patch tokens.

To model spatial information from multi-frames, one simple way is directly feeding all tokens into CLIP's vision encoder and conducting attention across all tokens. However, this method introduces significant conflicts with CLIP. As CLIP is pre-trained on image and text pairs, it has difficulty handling interactions of tokens between images/frames during training. We also verify it by experiments as Full Attention setting in Table \ref{tab:model}. Instead, we introduce a Video Proxy token to act as a proxy that helps each local patch perceive video-level temporal information. 

Before feeding into CLIP, we concatenate patch tokens with a set of learnable parameters called video proxy tokens: $\mathcal{P}=\{p_1, p_2, ..., p_M\}$, where $M$ is the number of video proxy tokens. Then all $T\times N+M$ tokens will be fed into the ViT of CLIP. The output of the first video proxy token will be regarded as the video's representation. We also design a proxy-guided attention mechanism for the vanilla ViT. In the attention score calculation of each block, video proxy tokens attend to all tokens, while patch tokens only attend to tokens in the same frame plus video proxy tokens. This mechanism can be formulated as attention mask $\mathcal{M}_{\mathrm{ViP}}$:
\begin{equation}
\begin{aligned}
\mathcal{M}_{\mathrm{ViP}}(q, k) = 
\begin{cases}
            1 & \text{if $q \in \mathcal{P}$ or $k \in \mathcal{P}$ or $(q, k)$ in the same frame,} \\
            0 & \text{otherwise,}
\end{cases}
\end{aligned}
\end{equation}
where $q$ and $k$ is the key and query in attention calculation. Patch tokens can obtain global information from video proxy tokens while reducing inconsistencies with the original CLIP's calculation. Our experiment in Section \ref{sec:exp} demonstrates the superiority of this mechanism.

For the input type of the image/frame, we use linear interpolation to get a middle temporal positional embedding, then treat the image/frame as a special single-frame video. This method enables joint training on both videos and images in the same batch, as our proxy-guided attention mechanism reduces the difference in calculations between video and image. 

% \subsection{Omnisource Joint Learning}
\vspace{-5pt}
\subsection{Omnisource Cross-Modal Learning} \label{sec:ocl}
\vspace{-5pt}
To learn rich video-language alignment from video-subtitle pairs and reduce the language domain gap with downstream data by corresponding auxiliary frame-caption pairs, we study joint Cross-Modal Learning on the omnisource input.
Following most works of learning multimodal alignment on dual encoders \citep{radford2021learning,xue2022advancing,Li2021AlignBF,luo2020univl,xu2021videoclip,luo2021clip4clip}, we use info-NCE loss to perform contrastive learning. There are two formats of visual source : video sequences and single frames, and two types of text source : subtitles and captions in our work. We denote them by $V$, $F$, $S$, and $C$ respectively. We define a source-wise info-NCE loss by:
% \begin{equation}
% \begin{aligned}
% &\mathcal{L}_{v 2 t} =-\frac{1}{B} \sum_{i=1}^{B} \log \frac{\exp \left(v_{i}^{\top} t_{i} / \tau\right)}{\sum_{j=1}^{B} \exp \left(v_{i}^{\top} t_{j} / \tau\right)} \\
% &\mathcal{L}_{t 2 v} =-\frac{1}{B} \sum_{i=1}^{B} \log \frac{\exp \left(t_{i}^{\top} v_{i} / \tau\right)}{\sum_{j=1}^{B} \exp \left(t_{i}^{\top} v_{j} / \tau\right)} \\
% &\mathcal{L}_{X \leftrightarrow Y} = \mathcal{L}_{v 2 t} + \mathcal{L}_{t 2 v},
% \end{aligned}
% \end{equation}

% \begin{equation}
% \small
% % \begin{aligned}
% \mathcal{L}_{v 2 t} =-\frac{1}{B} \sum_{i=1}^{B} \log \frac{\exp \left(v_{i}^{\top} t_{i} / \tau\right)}{\sum_{j=1}^{B} \exp \left(v_{i}^{\top} t_{j} / \tau\right)}, ~~~~
% \mathcal{L}_{t 2 v} =-\frac{1}{B} \sum_{i=1}^{B} \log \frac{\exp \left(t_{i}^{\top} v_{i} / \tau\right)}{\sum_{j=1}^{B} \exp \left(t_{i}^{\top} v_{j} / \tau\right)}
% % \end{aligned}
% \end{equation}
\begin{equation}
\mathcal{L}_{v 2 t} =-\frac{1}{B} \sum_{i=1}^{B} \log \frac{e^{v_{i}^{\top} t_{i} / \tau}}{\sum_{j=1}^{B} e^{v_{i}^{\top} t_{j} / \tau}}, ~~~~
\mathcal{L}_{t 2 v} =-\frac{1}{B} \sum_{i=1}^{B} \log \frac{e^{t_{i}^{\top} v_{i} / \tau}}{\sum_{j=1}^{B} e^{t_{i}^{\top} v_{j} / \tau}}
\end{equation}
where $v_i$ and $t_j$ are the normalized embeddings of $i$-th visual feature in $X \in \{V, F\}$ and $j$-th text feature in $Y \in \{S, C\}$ in a batch of size $B$. $\tau$ is a learnable temperature. The overall alignment loss $\mathcal{L}_{X \leftrightarrow Y}$ is the average of $\mathcal{L}_{v 2 t}$ and $\mathcal{L}_{t 2 v}$. For example, $\mathcal{L}_{V \leftrightarrow S}$ represents info-NCE loss within video-subtitle pairs in a batch, which pulls aligned pairs together in embedding space while pushing apart misaligned pairs. 

We study the reasonable variants of OCL: 
\textbf{(a)} $\mathcal{L}_{V \leftrightarrow S} + \mathcal{L}_{F \leftrightarrow C}$: Simple combination of two source-wise losses on video-subtitle and frame-caption pairs; 
\textbf{(b)} $\mathcal{L}_{V \leftrightarrow S} + \mathcal{L}_{V \leftrightarrow C}$: As there is also content correlation between videos and its middle-frame captions, we explore to add a loss on video-caption pairs to baseline loss $\mathcal{L}_{V \leftrightarrow S}$; 
\textbf{(c)} $\mathcal{L}_{V \leftrightarrow S} + \mathcal{L}_{V \leftrightarrow C} + \mathcal{L}_{F \leftrightarrow C}$: Combination of (a) and (c); 
\textbf{(d)} $\mathcal{L}_{V \leftrightarrow S,C} + \mathcal{L}_{F \leftrightarrow C}$: A video corresponds to both a subtitle and auxiliary caption. Compare to (c), the numbers of negative pairs in $\mathcal{L}_{v 2 t}$ can be expanded.  The $\mathcal{L}_{v 2 t}$ in $\mathcal{L}_{V \leftrightarrow S,C}$ is rewritten as:
% \begin{equation}
% \small
% \begin{aligned}
% \mathcal{L}_{v 2 t} =-\frac{1}{2B} \sum_{i=1}^{B} (\log \frac{\exp \left(v_{i}^{\top} s_{i} / \tau\right)}{\sum_{j=1}^{B} \exp \left(v_{i}^{\top} s_{j} / \tau\right) + \exp \left(v_{i}^{\top} c_{j \neq i} / \tau\right)} + \log \frac{\exp \left(v_{i}^{\top} c_{i} / \tau\right)}{\sum_{j=1}^{B} \exp \left(v_{i}^{\top} c_{j} / \tau\right) + \exp \left(v_{i}^{\top} s_{j \neq i} / \tau\right)}
% ),
% \end{aligned}
% \end{equation}
\begin{equation}
\begin{aligned}
\mathcal{L}_{v 2 t} =-\frac{1}{2B} \sum_{i=1}^{B} (\log \frac{e^{v_{i}^{\top} s_{i} / \tau}}{\sum_{j=1}^{B} e^{v_{i}^{\top} s_{j} / \tau} + e^{v_{i}^{\top} c_{j \neq i} / \tau}} + \log \frac{e^{v_{i}^{\top} c_{i} / \tau}}{\sum_{j=1}^{B} e^{v_{i}^{\top} c_{j} / \tau} + e^{v_{i}^{\top} s_{j \neq i} / \tau}}
),
\end{aligned}
\end{equation}
where $s_i \in S$ and $c_i \in C$. The $\mathcal{L}_{t 2 v}$ in $\mathcal{L}_{V \leftrightarrow S,C}$ is equal to \textbf{(c)}. We compare all variants with the baseline $\mathcal{L}_{V \leftrightarrow S}$ and report results in Section \ref{sec:exp}.

\section{Experiment}\label{sec:exp}

\tabablmodel
\tabablloss
\tababldata
\vspace{-8pt}

% We first describe the implementation 
% details of post-pretraining and finetuning. Then a thorough ablation study is conducted to demonstrate the effectiveness of our post-pretraining, from aspects of the model, data, and loss function. Finally, we make comparisons to existing state-of-the-art methods.
% In this section, we first describe the implementation details during video-text post-pretraining and fine-tuning downstream tasks. Next, we conduct ablation studies to demonstrate the effectiveness of our designed CLIP-ViP model which learn video-language representation by video and image joint pre-training. Finally, we make comparisons between our model and the state-of-the-art methods.

\subsection{Experimental Details}
% \paragraph{Post-pretraining details.} We uniformly sample 12 frames for each video and resize frame size to 224*224. For language, we adopt the CLIP's tokenizer to split a sentence into word tokens with a max length of 70. We use AdamW optimizer~\citep{loshchilov2017adamw} with an initial learning rate of 5e-6 and a fixed weight decay of 5e-2. For learning rate schedule, we adopt a cosine decay with a warm-up strategy. We train our model with 32 NVIDIA Tesla V100 GPUs in a batch size of 1024. The contrastive similarity is calculated on gathered features from all GPUs. We set training steps equal to one epoch on HD-VILA-100M for all ablation studies and 3 epochs for the full setting.
\paragraph{Video-Text Post-Pretraining.} To pre-train the proposed CLIP-ViP model, we uniformly sample 12 frames and resize all frames to 224$\times$224 from video clips with an average length of 13.4 seconds. The sampled frames can well cover the semantics conveyed from a video clip. For text, we adopt the CLIP's tokenizer to split a sentence into word tokens with a max length of 70. We use AdamW optimizer~\citep{loshchilov2017adamw}, and empirically set an initial learning rate as 5e-6 and a fixed weight decay as 5e-2. For the learning rate schedule, we adopt a cosine decay with a warm-up strategy. We train our model with 32 NVIDIA Tesla V100 GPUs in a batch size of 1024. The contrastive similarity is calculated on gathered features from all GPUs. We set training steps to one epoch on HD-VILA-100M for all ablation studies and three epochs for the full setting.

% \paragraph{Downstream details.} We reuse most hyper-parameters in post-pretraining, except for: 1) batch size, we finetune our model in a batch size of 128 for all downstream tasks. 2) learning rate and weight decay, we set lr and weight decay to 1e-6 and 0.2, respectively. 3) number of epochs, Due to the different sizes of downstream datasets, we set number of epochs as 5, 20, 10, 20 for MSR-VTT, DiDeMo, LSMDC, ActivityNet, respectively. 4) frame length, we set frame length to 32 for ActivityNet Captions as its videos are much longer (180 seconds on average). The hyper-parameters of downstream finetuning are the same in all ablation studies.
\paragraph{Fine-tuning Training.} To better adapt CLIP-ViP to downstream tasks, we reuse most hyper-parameters of post-pretraining in fine-tuning with some exceptions. 1) Batch size: we fine-tune our model with a batch size of 128 for all downstream tasks for a fair comparison. 2) Learning rate and weight decay: we empirically set them to 1e-6 and 0.2, respectively. 3) Number of epochs: due to the various scales of downstream datasets, we set epoch numbers to 5, 20, 10, and 20 for MSR-VTT, DiDeMo, LSMDC, and ActivityNet, respectively. 4) Frame number: for a fair comparison, we set frame number to 12 except for ActivityNet Captions (set to 32) as its videos are much longer (180 seconds on average). Note that the hyper-parameters of downstream training are the same in all settings in the ablation study.

% \paragraph{Downstream Datasets.} We conduct video-text retrieval experiments on four datasets. \textbf{(a) MSR-VTT}~\citep{xu2016msr} contains 10K YouTube videos with 200K descriptions. We follow previous works~\citep{yu2018jsfusion,liu2019use}, training models on 9K videos, and reporting results on the 1K-A test set. \textbf{(b) DiDeMo}~\citep{anne2017localizing} consists of 10K Flickr videos annotated with 40K sentences. We follow~\citep{liu2019use, zhang2018cross} to evaluate paragraph-to-video retrieval, where all descriptions for a video are concatenated to form a single query. \textbf{(c) LSMDC}~\citep{Rohrbach2016MovieD} consists of 118,081 video clips sourced from 202 movies. Each video has a caption. Evaluation is conducted on a test set of 1,000 videos from movies disjoint from the train and validation sets. \textbf{(d) ActivityNet Captions}~\citep{Krishna2017actnetcaption} contains 20K YouTube videos annotated with 100K sentences. We follow the paragraph-to-video retrieval protocols \citep{zhang2018cross,liu2019use} training on 10K videos and reporting results on the val1 set with 4.9K videos.
\paragraph{Downstream Datasets.} To evaluate performances of video pre-training models, we conduct text-to-video retrieval experiments on four typical datasets. \textbf{(a) MSR-VTT}~\citep{xu2016msr} contains 10K YouTube videos with 200K descriptions. We follow previous works~\citep{yu2018jsfusion,liu2019use} to train models on 9K videos, and report results on the 1K-A test set. \textbf{(b) DiDeMo}~\citep{anne2017localizing} consists of 10K Flickr videos annotated with 40K sentences. We follow~\citep{liu2019use, zhang2018cross} to evaluate paragraph-to-video retrieval and concatenate all descriptions of a video as one query. \textbf{(c) LSMDC}~\citep{Rohrbach2016MovieD} consists of 118,081 video clips sourced from 202 movies with one caption corresponding to each clip. Evaluation is conducted on a test set of 1,000 videos from movies disjoint from the train and validation sets. \textbf{(d) ActivityNet Captions}~\citep{Krishna2017actnetcaption} contains 20K YouTube videos annotated with 100K sentences. We follow the paragraph-to-video retrieval setting \citep{zhang2018cross,liu2019use} to train models on 10K videos and report results on the val1 set with 4.9K videos.

\subsection{Ablation Studies}\label{sec:abl}
% \paragraph{Vision Encoder.}
\paragraph{Video Proxy Mechanism.}

% We evaluate different model structures: MeanPool, SeqTransformer, Full Attention, Different number of video proxies, by finetuning on MSR-VTT. MeanPool simply takes the average of frame features as the representation of the whole video. For SeqTransformer, we follow the seqTransf type in \citep{luo2021clip4clip} and the residual connection in their implementation. Full attention type takes all patch tokens as the input of the vision encoder and Attention calculation is on all tokens. We also study the impact of different numbers of video proxies. All models are initialized from a CLIP-ViT-B/32. The results are shown in Table \ref{tab:model}. The simplest model is MeanPool type, which completely disregard temporality. Compare to the MeanPool baseline, SeqTransformer improves the average Recall@1,5,10 by 0.8\%. Full Attention type leads to a significant performance drop, although this type allows individual features to interact globally. Besides, we observe the type has a worse initial status and converges slower than other types. One explanation is full Attention has a significant calculation conflict with CLIP's in-frame Attention. From Table \ref{tab:model}, our Video-Proxy mechanism has the most improvement. Different numbers of Video proxies all result in significant performance gains on R@1 (3.1\% by 4 proxies), while only increasing negligible parameters and computation cost.
For the vision encoder, we evaluate our proposed Video Proxy (ViP) mechanism with different numbers of proxies and compare it with different model structures (i.e. MeanPool, SeqTransformer, Full Attention) by fine-tuning the same pre-trained model on MSR-VTT retrieval task. MeanPool simply takes the average of frame features as the representation of the whole video. For SeqTransformer, we follow the seqTransf type in CLIP4Clip \citep{luo2021clip4clip} and the residual connection in their implementation. Full Attention setting takes all patch tokens as the input of the vision encoder and attention is conducted across all tokens. All models are initialized with CLIP-ViT-B/32. The results are shown in Table \ref{tab:model}. Compared to the MeanPool baseline which completely disregards temporality, SeqTransformer improves the average Recall@1,5,10 by 0.8\%. Full Attention type leads to a significant performance drop and we observe a worse initial status and slower convergence than other settings during the experiment. This is consistent with our analysis in Section \ref{sec:vip} that directly using CLIP for all patches' attention computation will decrease the advantage of CLIP. In our method, different numbers of video proxy tokens all result in significant performance gain on R@1 (e.g., 3.1\% by 4 proxies), while only increasing negligible parameters: 3K compared to 86M of ViT backbone. Compared with other settings, our methods in all settings have the most improvement which indicates that our proposed video proxy mechanism can effectively leverage the image-text pre-trained model for video-text post-pretraining.

\paragraph{Omnisource Cross-modal Learning.}
To verify the effectiveness of the proposed Omnisource Cross-modal Learning (OCL) and compare its variants, we set a post-pretraining and fine-tuning pipeline and adopt the same hyper-parameters for all experiments. $\mathcal{L}_{V \leftrightarrow S}$ is the baseline contrastive loss on video-subtitle pairs. After introducing auxiliary captions, we study four variants of OCL Loss: 
(a) $\mathcal{L}_{V \leftrightarrow S} + \mathcal{L}_{F \leftrightarrow C}$; 
(b) $\mathcal{L}_{V \leftrightarrow S} + \mathcal{L}_{V \leftrightarrow C}$; 
(c) $\mathcal{L}_{V \leftrightarrow S} + \mathcal{L}_{V \leftrightarrow C} + \mathcal{L}_{F \leftrightarrow C}$; 
(d) $\mathcal{L}_{V \leftrightarrow S,C} + \mathcal{L}_{F \leftrightarrow C}$ as explained in Section \ref{sec:ocl}. We pre-train models with each loss function for only one epoch due to the costly training, then finetune on two video-text retrieval datasets: MSR-VTT and DiDeMo. 
We compare the results with CLIP-MeanPool and CLIP using the proposed Video Proxy mechanism without post-pretraining (i.e., CLIP-ViP). The results are listed in Table \ref{tab:loss}. 
On MSR-VTT dataset, we find that $\mathcal{L}_{V \leftrightarrow S}$ brings very little improvement: 0.4\% on average of Recall@1, 5, 10. This is due to the large domain gap between MSR-VTT and post-pretraining data. Combined with auxiliary captions, four variants of OCL loss all bring significant improvements: over 3\% on Recall@1 and over 2.3\% on average of Recall@1, 5, 10. On DiDeMo dataset, based on the improvement brought by $\mathcal{L}_{V \leftrightarrow S}$, OCL further improve the results by a large margin: 8\% on average of Recall@1, 5, 10. Finally, $\mathcal{L}_{V \leftrightarrow S,C} + \mathcal{L}_{F \leftrightarrow C}$ performs best which is applied as our final setting. 

\tabmsrvttactnet
\vspace{-8pt}

\paragraph{Auxiliary Data.}
In this part, we ablate the contribution of large-scale noisy data and auxiliary data. For uni-source, we use video-subtitle pairs and video-caption data for post-pretraining by vanilla contrastive loss. For data combination, we apply OCL under $\mathcal{L}_{V \leftrightarrow S,C} + \mathcal{L}_{F \leftrightarrow C}$ setting to post-pretrain on the combined data. From Table \ref{tab:data}, Omnisource post-pretraining results are much better than two uni-source results. On MSR-VTT, both uni-source post-pretraining show limited improvement: 67.4\% and 66.9\% compared with 67.0\%. While the Omnisource post-pretraining brings a significant improvement of 2.6\%. On DiDeMo, the benefit of data combination is also obvious, with nearly double the improvements brought by uni-source. These results show that the auxiliary data together with our designed joint learning method can effectively adapt the image-text pre-trained model to video-text post-pretraining.

As the generation of auxiliary captions is based on OFA-Caption \citep{wang2022unifying}, a powerful image-text pre-trained model, we also explore only including existing data in post-pretraining. We choose image-text pairs of several widely-adopted datasets: MS-COCO, Visual Genome (VG) \citep{krishna2017visual}, Flickr-30K \citep{young2014image}, SBU \citep{ordonez2011im2text}, CC3M \citep{sharma2018conceptual} and CC12M as our auxiliary data (namely ImageCaption). To ablate the contribution of these data, we add experiments of post-pretraining on ImageCaption alone and HD-VILA-100M combined with ImageCaption. From Table \ref{tab:data}, post-pretraining on ImageCaption alone results in performance degradation on MSR-VTT and marginal improvement on DiDeMo. In contrast, ImageCaption yields significant performance gains on both datasets when used as auxiliary data for HD-VILA-100M. This further illustrates the importance of the combination of large-scale noisy data and auxiliary data.

\subsection{Comparison to State-of-the-art Models}

\tabDiDeMolsmdc
\vspace{-5pt}
We compare our model under full setting (in three epochs) with state-of-the-art works on the text-to-video retrieval task. The results of fine-tuning on four datasets (i.e., MSR-VTT, DiDeMo, ActivityNet Captions, LSMDC) are shown in Table \ref{tab:retrieval-msrvttactnet} and \ref{tab:retrieval-DiDeMolsmdc}, respectively. We clarify the backbone for CLIP-based works. We only add results with DSL \citep{cheng2021improving} to make fair comparison with some methods using post-processing operations e.g., DSL \citep{cheng2021improving} or QB-Norm \citep{bogolin2022cross}.
Our model achieves the best results on all datasets in both CLIP-ViT-B/32 and CLIP-ViT-B/16. Note that some existing methods are also applicable on top of our models as our modification to the CLIP model is minimal. 
% For MSR-VTT, DiDeMo, and the ActivityNet Captions dataset, our model outperforms existing methods by a large margin in both CLIP-ViT-B/32 and CLIP-ViT-B/16. For LSMDC, our model achieves the best results on CLIP-ViT-B/32 and by a large margin on CLIP-ViT-B/16. 
Note that even without post-processing (e.g., DSL), our results still surpass methods using post-processing operations on most datasets. Besides, adding DSL will greatly improve the performance of our model since our model has good bidirectional vision-language correspondence. The good results on the ActivityNet Captions dataset also indicate that our model can generalize well to long videos. Overall, the improvements on different datasets demonstrate the superiority of the video-language representation learned by our CLIP-ViP model.

\vspace{-1mm}
\section{Conclusion}
\vspace{-1mm}
In this paper, we study further pre-training (post-pretraining) image-text models like CLIP on large-scale video data. We first conduct a preliminary analysis to reveal the factors hindering video post-pretraining. Motivated by findings, we propose CLIP-ViP which includes an Omnisource Cross-modal Learning method and a Video Proxy mechanism. The Video Proxy mechanism can better model videos containing temporal information while reducing conflicts with the pre-trained CLIP model.
The Omnisource Cross-modal Learning alleviates the problem caused by the domain gap between video-subtitle and downstream data.
Extensive results show that our approach improves the performance of CLIP on video-text retrieval by a large margin and also achieves new state-of-the-art results on four widely-used video-language benchmarks.

% \paragraph{Acknowledgement}

\bibliography{iclr2023_conference}

\begin{thebibliography}{63}
\providecommand{\natexlab}[1]{#1}
\providecommand{\url}[1]{\texttt{#1}}
\expandafter\ifx\csname urlstyle\endcsname\relax
  \providecommand{\doi}[1]{doi: #1}\else
  \providecommand{\doi}{doi: \begingroup \urlstyle{rm}\Url}\fi

\bibitem[Anne~Hendricks et~al.(2017)Anne~Hendricks, Wang, Shechtman, Sivic,
  Darrell, and Russell]{anne2017localizing}
Lisa Anne~Hendricks, Oliver Wang, Eli Shechtman, Josef Sivic, Trevor Darrell,
  and Bryan Russell.
\newblock Localizing moments in video with natural language.
\newblock In \emph{ICCV}, pp.\  5803--5812, 2017.

\bibitem[Bain et~al.(2021)Bain, Nagrani, Varol, and Zisserman]{bain2021frozen}
Max Bain, Arsha Nagrani, G{\"u}l Varol, and Andrew Zisserman.
\newblock Frozen in time: A joint video and image encoder for end-to-end
  retrieval.
\newblock In \emph{ICCV}, 2021.

\bibitem[Bogolin et~al.(2022)Bogolin, Croitoru, Jin, Liu, and
  Albanie]{bogolin2022cross}
Simion-Vlad Bogolin, Ioana Croitoru, Hailin Jin, Yang Liu, and Samuel Albanie.
\newblock Cross modal retrieval with querybank normalisation.
\newblock In \emph{CVPR}, pp.\  5194--5205, 2022.

\bibitem[Changpinyo et~al.(2021)Changpinyo, Sharma, Ding, and
  Soricut]{changpinyo2021conceptual}
Soravit Changpinyo, Piyush Sharma, Nan Ding, and Radu Soricut.
\newblock Conceptual 12m: Pushing web-scale image-text pre-training to
  recognize long-tail visual concepts.
\newblock In \emph{CVPR}, pp.\  3558--3568, 2021.

\bibitem[Chen et~al.(2020)Chen, Li, Yu, El~Kholy, Ahmed, Gan, Cheng, and
  Liu]{chen2020uniter}
Yen-Chun Chen, Linjie Li, Licheng Yu, Ahmed El~Kholy, Faisal Ahmed, Zhe Gan,
  Yu~Cheng, and Jingjing Liu.
\newblock Uniter: Universal image-text representation learning.
\newblock In \emph{ECCV}, pp.\  104--120, 2020.

\bibitem[Cheng et~al.(2021)Cheng, Lin, Wu, Yang, and Shen]{cheng2021improving}
Xing Cheng, Hezheng Lin, Xiangyu Wu, Fan Yang, and Dong Shen.
\newblock Improving video-text retrieval by multi-stream corpus alignment and
  dual softmax loss.
\newblock \emph{arXiv preprint arXiv:2109.04290}, 2021.

\bibitem[Fang et~al.(2021)Fang, Xiong, Xu, and Chen]{fang2021clip2video}
Han Fang, Pengfei Xiong, Luhui Xu, and Yu~Chen.
\newblock Clip2video: Mastering video-text retrieval via image clip.
\newblock \emph{arXiv preprint arXiv:2106.11097}, 2021.

\bibitem[Fu et~al.(2021)Fu, Li, Gan, Lin, Wang, Wang, and Liu]{fu2021violet}
Tsu-Jui Fu, Linjie Li, Zhe Gan, Kevin Lin, William~Yang Wang, Lijuan Wang, and
  Zicheng Liu.
\newblock Violet: End-to-end video-language transformers with masked
  visual-token modeling.
\newblock \emph{arXiv preprint arXiv:2111.12681}, 2021.

\bibitem[Gabeur et~al.(2020)Gabeur, Sun, Alahari, and Schmid]{gabeur2020multi}
Valentin Gabeur, Chen Sun, Karteek Alahari, and Cordelia Schmid.
\newblock Multi-modal transformer for video retrieval.
\newblock In \emph{ECCV}, pp.\  214--229, 2020.

\bibitem[Gao et~al.(2021)Gao, Liu, Chen, Chang, Zhang, and
  Yuan]{gao2021clip2tv}
Zijian Gao, Jingyu Liu, Sheng Chen, Dedan Chang, Hao Zhang, and Jinwei Yuan.
\newblock Clip2tv: An empirical study on transformer-based methods for
  video-text retrieval.
\newblock \emph{arXiv preprint arXiv:2111.05610}, 2021.

\bibitem[Ge et~al.(2022)Ge, Ge, Liu, Li, Shan, Qie, and Luo]{ge2022bridging}
Yuying Ge, Yixiao Ge, Xihui Liu, Dian Li, Ying Shan, Xiaohu Qie, and Ping Luo.
\newblock Bridging video-text retrieval with multiple choice questions.
\newblock In \emph{CVPR}, pp.\  16167--16176, 2022.

\bibitem[Gorti et~al.(2022)Gorti, Vouitsis, Ma, Golestan, Volkovs, Garg, and
  Yu]{gorti2022x}
Satya~Krishna Gorti, No{\"e}l Vouitsis, Junwei Ma, Keyvan Golestan, Maksims
  Volkovs, Animesh Garg, and Guangwei Yu.
\newblock X-pool: Cross-modal language-video attention for text-video
  retrieval.
\newblock In \emph{CVPR}, pp.\  5006--5015, 2022.

\bibitem[Goyal et~al.(2017)Goyal, Ebrahimi~Kahou, Michalski, Materzynska,
  Westphal, Kim, Haenel, Fruend, Yianilos, Mueller-Freitag,
  et~al.]{goyal2017something}
Raghav Goyal, Samira Ebrahimi~Kahou, Vincent Michalski, Joanna Materzynska,
  Susanne Westphal, Heuna Kim, Valentin Haenel, Ingo Fruend, Peter Yianilos,
  Moritz Mueller-Freitag, et~al.
\newblock The" something something" video database for learning and evaluating
  visual common sense.
\newblock In \emph{ICCV}, pp.\  5842--5850, 2017.

\bibitem[Gu et~al.(2021)Gu, Lin, Kuo, and Cui]{gu2021open}
Xiuye Gu, Tsung-Yi Lin, Weicheng Kuo, and Yin Cui.
\newblock Open-vocabulary object detection via vision and language knowledge
  distillation.
\newblock In \emph{ICLR}, 2021.

\bibitem[Guzhov et~al.(2022)Guzhov, Raue, Hees, and
  Dengel]{guzhov2022audioclip}
Andrey Guzhov, Federico Raue, J{\"o}rn Hees, and Andreas Dengel.
\newblock Audioclip: Extending clip to image, text and audio.
\newblock In \emph{ICASSP}, pp.\  976--980. IEEE, 2022.

\bibitem[Huang et~al.(2021{\natexlab{a}})Huang, Xue, Liu, and
  Lu]{huang2021unifying}
Yupan Huang, Hongwei Xue, Bei Liu, and Yutong Lu.
\newblock Unifying multimodal transformer for bi-directional image and text
  generation.
\newblock In \emph{ACM MM}, pp.\  1138--1147, 2021{\natexlab{a}}.

\bibitem[Huang et~al.(2020)Huang, Zeng, Liu, Fu, and Fu]{huang2020pixel}
Zhicheng Huang, Zhaoyang Zeng, Bei Liu, Dongmei Fu, and Jianlong Fu.
\newblock Pixel-bert: Aligning image pixels with text by deep multi-modal
  transformers.
\newblock \emph{arXiv preprint arXiv:2004.00849}, 2020.

\bibitem[Huang et~al.(2021{\natexlab{b}})Huang, Zeng, Huang, Liu, Fu, and
  Fu]{huang2021seeing}
Zhicheng Huang, Zhaoyang Zeng, Yupan Huang, Bei Liu, Dongmei Fu, and Jianlong
  Fu.
\newblock Seeing out of the box: End-to-end pre-training for vision-language
  representation learning.
\newblock In \emph{CVPR}, pp.\  12976--12985, 2021{\natexlab{b}}.

\bibitem[Jia et~al.(2021)Jia, Yang, Xia, Chen, Parekh, Pham, Le, Sung, Li, and
  Duerig]{jia2021scaling}
Chao Jia, Yinfei Yang, Ye~Xia, Yi-Ting Chen, Zarana Parekh, Hieu Pham, Quoc Le,
  Yun-Hsuan Sung, Zhen Li, and Tom Duerig.
\newblock Scaling up visual and vision-language representation learning with
  noisy text supervision.
\newblock In \emph{ICML}, pp.\  4904--4916. PMLR, 2021.

\bibitem[Kim et~al.(2021)Kim, Son, and Kim]{kim2021vilt}
Wonjae Kim, Bokyung Son, and Ildoo Kim.
\newblock Vilt: Vision-and-language transformer without convolution or region
  supervision.
\newblock In \emph{ICML}, pp.\  5583--5594. PMLR, 2021.

\bibitem[Krishna et~al.(2017{\natexlab{a}})Krishna, Hata, Ren, Fei-Fei, and
  Niebles]{Krishna2017actnetcaption}
Ranjay Krishna, Kenji Hata, Frederic Ren, Li~Fei-Fei, and Juan~Carlos Niebles.
\newblock Dense-captioning events in videos.
\newblock In \emph{ICCV}, pp.\  706--715, 2017{\natexlab{a}}.

\bibitem[Krishna et~al.(2017{\natexlab{b}})Krishna, Zhu, Groth, Johnson, Hata,
  Kravitz, Chen, Kalantidis, Li, Shamma, et~al.]{krishna2017visual}
Ranjay Krishna, Yuke Zhu, Oliver Groth, Justin Johnson, Kenji Hata, Joshua
  Kravitz, Stephanie Chen, Yannis Kalantidis, Li-Jia Li, David~A Shamma, et~al.
\newblock Visual genome: Connecting language and vision using crowdsourced
  dense image annotations.
\newblock \emph{IJCV}, 123\penalty0 (1):\penalty0 32--73, 2017{\natexlab{b}}.

\bibitem[Kuehne et~al.(2011)Kuehne, Jhuang, Garrote, Poggio, and
  Serre]{kuehne2011hmdb}
Hildegard Kuehne, Hueihan Jhuang, Est{\'\i}baliz Garrote, Tomaso Poggio, and
  Thomas Serre.
\newblock Hmdb: a large video database for human motion recognition.
\newblock In \emph{ICCV}, 2011.

\bibitem[Lei et~al.(2021)Lei, Li, Zhou, Gan, Berg, Bansal, and
  Liu]{lei2021less}
Jie Lei, Linjie Li, Luowei Zhou, Zhe Gan, Tamara~L Berg, Mohit Bansal, and
  Jingjing Liu.
\newblock Less is more: Clipbert for video-and-language learning via sparse
  sampling.
\newblock In \emph{CVPR}, pp.\  7331--7341, 2021.

\bibitem[Li et~al.(2020{\natexlab{a}})Li, Duan, Fang, Gong, and
  Jiang]{li2020unicoder}
Gen Li, Nan Duan, Yuejian Fang, Ming Gong, and Daxin Jiang.
\newblock Unicoder-vl: A universal encoder for vision and language by
  cross-modal pre-training.
\newblock In \emph{AAAI}, pp.\  11336--11344, 2020{\natexlab{a}}.

\bibitem[Li et~al.(2021)Li, Selvaraju, Gotmare, Joty, Xiong, and
  Hoi]{Li2021AlignBF}
Junnan Li, Ramprasaath~R. Selvaraju, Akhilesh~Deepak Gotmare, Shafiq~R. Joty,
  Caiming Xiong, and Steven C.~H. Hoi.
\newblock Align before fuse: Vision and language representation learning with
  momentum distillation.
\newblock In \emph{NeurIPS}, 2021.

\bibitem[Li et~al.(2020{\natexlab{b}})Li, Yin, Li, Zhang, Hu, Zhang, Wang, Hu,
  Dong, Wei, et~al.]{li2020oscar}
Xiujun Li, Xi~Yin, Chunyuan Li, Pengchuan Zhang, Xiaowei Hu, Lei Zhang, Lijuan
  Wang, Houdong Hu, Li~Dong, Furu Wei, et~al.
\newblock Oscar: Object-semantics aligned pre-training for vision-language
  tasks.
\newblock In \emph{ECCV}, pp.\  121--137, 2020{\natexlab{b}}.

\bibitem[Lin et~al.(2014)Lin, Maire, Belongie, Hays, Perona, Ramanan,
  Doll{\'a}r, and Zitnick]{lin2014microsoft}
Tsung-Yi Lin, Michael Maire, Serge Belongie, James Hays, Pietro Perona, Deva
  Ramanan, Piotr Doll{\'a}r, and C~Lawrence Zitnick.
\newblock Microsoft coco: Common objects in context.
\newblock In \emph{ECCV}, pp.\  740--755. Springer, 2014.

\bibitem[Liu et~al.(2019)Liu, Albanie, Nagrani, and Zisserman]{liu2019use}
Yang Liu, Samuel Albanie, Arsha Nagrani, and Andrew Zisserman.
\newblock Use what you have: Video retrieval using representations from
  collaborative experts.
\newblock In \emph{BMVC}, 2019.

\bibitem[Loshchilov \& Hutter(2019)Loshchilov and Hutter]{loshchilov2017adamw}
Ilya Loshchilov and Frank Hutter.
\newblock Decoupled weight decay regularization.
\newblock In \emph{ICLR}, 2019.

\bibitem[Luo et~al.(2020)Luo, Ji, Shi, Huang, Duan, Li, Li, Bharti, and
  Zhou]{luo2020univl}
Huaishao Luo, Lei Ji, Botian Shi, Haoyang Huang, Nan Duan, Tianrui Li, Jason
  Li, Taroon Bharti, and Ming Zhou.
\newblock Univl: A unified video and language pre-training model for multimodal
  understanding and generation.
\newblock \emph{arXiv preprint arXiv:2002.06353}, 2020.

\bibitem[Luo et~al.(2021)Luo, Ji, Zhong, Chen, Lei, Duan, and
  Li]{luo2021clip4clip}
Huaishao Luo, Lei Ji, Ming Zhong, Yang Chen, Wen Lei, Nan Duan, and Tianrui Li.
\newblock Clip4clip: An empirical study of clip for end to end video clip
  retrieval.
\newblock \emph{arXiv preprint arXiv:2104.08860}, 2021.

\bibitem[Miech et~al.(2019)Miech, Zhukov, Alayrac, Tapaswi, Laptev, and
  Sivic]{miech2019howto100m}
Antoine Miech, Dimitri Zhukov, Jean-Baptiste Alayrac, Makarand Tapaswi, Ivan
  Laptev, and Josef Sivic.
\newblock Howto100m: Learning a text-video embedding by watching hundred
  million narrated video clips.
\newblock In \emph{ICCV}, pp.\  2630--2640, 2019.

\bibitem[Mokady et~al.(2021)Mokady, Hertz, and Bermano]{mokady2021clipcap}
Ron Mokady, Amir Hertz, and Amit~H Bermano.
\newblock Clipcap: Clip prefix for image captioning.
\newblock \emph{arXiv preprint arXiv:2111.09734}, 2021.

\bibitem[Ordonez et~al.(2011)Ordonez, Kulkarni, and Berg]{ordonez2011im2text}
Vicente Ordonez, Girish Kulkarni, and Tamara Berg.
\newblock Im2text: Describing images using 1 million captioned photographs.
\newblock \emph{NeurIPS}, 24, 2011.

\bibitem[Patashnik et~al.(2021)Patashnik, Wu, Shechtman, Cohen-Or, and
  Lischinski]{patashnik2021styleclip}
Or~Patashnik, Zongze Wu, Eli Shechtman, Daniel Cohen-Or, and Dani Lischinski.
\newblock Styleclip: Text-driven manipulation of stylegan imagery.
\newblock In \emph{ICCV}, pp.\  2085--2094, 2021.

\bibitem[Patrick et~al.(2021)Patrick, Huang, Asano, Metze, Hauptmann,
  Henriques, and Vedaldi]{patrick2021supportset}
Mandela Patrick, Po-Yao Huang, Yuki Asano, Florian Metze, Alexander~G
  Hauptmann, Joao~F. Henriques, and Andrea Vedaldi.
\newblock Support-set bottlenecks for video-text representation learning.
\newblock In \emph{ICLR}, 2021.

\bibitem[Radford et~al.(2021)Radford, Kim, Hallacy, Ramesh, Goh, Agarwal,
  Sastry, Askell, Mishkin, Clark, et~al.]{radford2021learning}
Alec Radford, Jong~Wook Kim, Chris Hallacy, Aditya Ramesh, Gabriel Goh,
  Sandhini Agarwal, Girish Sastry, Amanda Askell, Pamela Mishkin, Jack Clark,
  et~al.
\newblock Learning transferable visual models from natural language
  supervision.
\newblock In \emph{ICML}, pp.\  8748--8763. PMLR, 2021.

\bibitem[Rao et~al.(2022)Rao, Zhao, Chen, Tang, Zhu, Huang, Zhou, and
  Lu]{rao2022denseclip}
Yongming Rao, Wenliang Zhao, Guangyi Chen, Yansong Tang, Zheng Zhu, Guan Huang,
  Jie Zhou, and Jiwen Lu.
\newblock Denseclip: Language-guided dense prediction with context-aware
  prompting.
\newblock In \emph{CVPR}, pp.\  18082--18091, 2022.

\bibitem[Rohrbach et~al.(2016)Rohrbach, Torabi, Rohrbach, Tandon, Pal,
  Larochelle, Courville, and Schiele]{Rohrbach2016MovieD}
Anna Rohrbach, Atousa Torabi, Marcus Rohrbach, Niket Tandon, Christopher~Joseph
  Pal, H.~Larochelle, Aaron~C. Courville, and Bernt Schiele.
\newblock Movie description.
\newblock \emph{IJCV}, pp.\  94--120, 2016.

\bibitem[Sharma et~al.(2018)Sharma, Ding, Goodman, and
  Soricut]{sharma2018conceptual}
Piyush Sharma, Nan Ding, Sebastian Goodman, and Radu Soricut.
\newblock Conceptual captions: A cleaned, hypernymed, image alt-text dataset
  for automatic image captioning.
\newblock In \emph{ACL}, 2018.

\bibitem[Sun et~al.(2019)Sun, Myers, Vondrick, Murphy, and
  Schmid]{sun2019videobert}
Chen Sun, Austin Myers, Carl Vondrick, Kevin Murphy, and Cordelia Schmid.
\newblock Videobert: A joint model for video and language representation
  learning.
\newblock In \emph{ICCV}, pp.\  7464--7473, 2019.

\bibitem[Sun et~al.(2022)Sun, Xue, Song, Liu, Yang, and Fu]{sun2022long}
Yuchong Sun, Hongwei Xue, Ruihua Song, Bei Liu, Huan Yang, and Jianlong Fu.
\newblock Long-form video-language pre-training with multimodal temporal
  contrastive learning.
\newblock In \emph{NeurIPS}, 2022.

\bibitem[Wang et~al.(2022{\natexlab{a}})Wang, Ge, Yan, Ge, Lin, Cai, Wu, Shan,
  Qie, and Shou]{wang2022all}
Alex~Jinpeng Wang, Yixiao Ge, Rui Yan, Yuying Ge, Xudong Lin, Guanyu Cai,
  Jianping Wu, Ying Shan, Xiaohu Qie, and Mike~Zheng Shou.
\newblock All in one: Exploring unified video-language pre-training.
\newblock \emph{arXiv preprint arXiv:2203.07303}, 2022{\natexlab{a}}.

\bibitem[Wang et~al.(2021{\natexlab{a}})Wang, Xing, and
  Liu]{wang2021actionclip}
Mengmeng Wang, Jiazheng Xing, and Yong Liu.
\newblock Actionclip: A new paradigm for video action recognition.
\newblock \emph{arXiv preprint arXiv:2109.08472}, 2021{\natexlab{a}}.

\bibitem[Wang et~al.(2022{\natexlab{b}})Wang, Yang, Men, Lin, Bai, Li, Ma,
  Zhou, Zhou, and Yang]{wang2022unifying}
Peng Wang, An~Yang, Rui Men, Junyang Lin, Shuai Bai, Zhikang Li, Jianxin Ma,
  Chang Zhou, Jingren Zhou, and Hongxia Yang.
\newblock Unifying architectures, tasks, and modalities through a simple
  sequence-to-sequence learning framework.
\newblock \emph{arXiv preprint arXiv:2202.03052}, 2022{\natexlab{b}}.

\bibitem[Wang et~al.(2022{\natexlab{c}})Wang, Zhang, Zheng, Pan, and
  Hua]{wang2022disentangled}
Qiang Wang, Yanhao Zhang, Yun Zheng, Pan Pan, and Xian-Sheng Hua.
\newblock Disentangled representation learning for text-video retrieval.
\newblock \emph{arXiv preprint arXiv:2203.07111}, 2022{\natexlab{c}}.

\bibitem[Wang et~al.(2022{\natexlab{d}})Wang, Chen, Wu, Chen, Dai, Liu, Jiang,
  Zhou, and Yuan]{wang2022bevt}
Rui Wang, Dongdong Chen, Zuxuan Wu, Yinpeng Chen, Xiyang Dai, Mengchen Liu,
  Yu-Gang Jiang, Luowei Zhou, and Lu~Yuan.
\newblock Bevt: Bert pretraining of video transformers.
\newblock In \emph{CVPR}, pp.\  14733--14743, 2022{\natexlab{d}}.

\bibitem[Wang et~al.(2021{\natexlab{b}})Wang, Yu, Yu, Dai, Tsvetkov, and
  Cao]{wang2021simvlm}
Zirui Wang, Jiahui Yu, Adams~Wei Yu, Zihang Dai, Yulia Tsvetkov, and Yuan Cao.
\newblock Simvlm: Simple visual language model pretraining with weak
  supervision.
\newblock In \emph{ICLR}, 2021{\natexlab{b}}.

\bibitem[Xu et~al.(2021{\natexlab{a}})Xu, Ghosh, Huang, Arora, Aminzadeh,
  Feichtenhofer, Metze, and Zettlemoyer]{xu2021vlm}
Hu~Xu, Gargi Ghosh, Po-Yao Huang, Prahal Arora, Masoumeh Aminzadeh, Christoph
  Feichtenhofer, Florian Metze, and Luke Zettlemoyer.
\newblock Vlm: Task-agnostic video-language model pre-training for video
  understanding.
\newblock \emph{arXiv preprint arXiv:2105.09996}, 2021{\natexlab{a}}.

\bibitem[Xu et~al.(2021{\natexlab{b}})Xu, Ghosh, Huang, Okhonko, Aghajanyan,
  Metze, Zettlemoyer, and Feichtenhofer]{xu2021videoclip}
Hu~Xu, Gargi Ghosh, Po-Yao Huang, Dmytro Okhonko, Armen Aghajanyan, Florian
  Metze, Luke Zettlemoyer, and Christoph Feichtenhofer.
\newblock Videoclip: Contrastive pre-training for zero-shot video-text
  understanding.
\newblock In \emph{EMNLP}, pp.\  6787--6800, 2021{\natexlab{b}}.

\bibitem[Xu et~al.(2016)Xu, Mei, Yao, and Rui]{xu2016msr}
Jun Xu, Tao Mei, Ting Yao, and Yong Rui.
\newblock Msr-vtt: A large video description dataset for bridging video and
  language.
\newblock In \emph{CVPR}, pp.\  5288--5296, 2016.

\bibitem[Xue et~al.(2021)Xue, Huang, Liu, Peng, Fu, Li, and
  Luo]{xue2021probing}
Hongwei Xue, Yupan Huang, Bei Liu, Houwen Peng, Jianlong Fu, Houqiang Li, and
  Jiebo Luo.
\newblock Probing inter-modality: Visual parsing with self-attention for
  vision-and-language pre-training.
\newblock In \emph{NeurIPS}, 2021.

\bibitem[Xue et~al.(2022)Xue, Hang, Zeng, Sun, Liu, Yang, Fu, and
  Guo]{xue2022advancing}
Hongwei Xue, Tiankai Hang, Yanhong Zeng, Yuchong Sun, Bei Liu, Huan Yang,
  Jianlong Fu, and Baining Guo.
\newblock Advancing high-resolution video-language representation with
  large-scale video transcriptions.
\newblock In \emph{CVPR}, pp.\  5036--5045, 2022.

\bibitem[Young et~al.(2014)Young, Lai, Hodosh, and Hockenmaier]{young2014image}
Peter Young, Alice Lai, Micah Hodosh, and Julia Hockenmaier.
\newblock From image descriptions to visual denotations: New similarity metrics
  for semantic inference over event descriptions.
\newblock \emph{ACL}, 2:\penalty0 67--78, 2014.

\bibitem[Yu et~al.(2018)Yu, Kim, and Kim]{yu2018jsfusion}
Youngjae Yu, Jongseok Kim, and Gunhee Kim.
\newblock A joint sequence fusion model for video question answering and
  retrieval.
\newblock In \emph{ECCV}, pp.\  471--487, 2018.

\bibitem[Yuan et~al.(2021)Yuan, Chen, Chen, Codella, Dai, Gao, Hu, Huang, Li,
  Li, et~al.]{yuan2021florence}
Lu~Yuan, Dongdong Chen, Yi-Ling Chen, Noel Codella, Xiyang Dai, Jianfeng Gao,
  Houdong Hu, Xuedong Huang, Boxin Li, Chunyuan Li, et~al.
\newblock Florence: A new foundation model for computer vision.
\newblock \emph{arXiv preprint arXiv:2111.11432}, 2021.

\bibitem[Zellers et~al.(2021)Zellers, Lu, Hessel, Yu, Park, Cao, Farhadi, and
  Choi]{zellers2021merlot}
Rowan Zellers, Ximing Lu, Jack Hessel, Youngjae Yu, Jae~Sung Park, Jize Cao,
  Ali Farhadi, and Yejin Choi.
\newblock Merlot: Multimodal neural script knowledge models.
\newblock In \emph{NeurIPS}, 2021.

\bibitem[Zellers et~al.(2022)Zellers, Lu, Lu, Yu, Zhao, Salehi, Kusupati,
  Hessel, Farhadi, and Choi]{zellers2022merlot}
Rowan Zellers, Jiasen Lu, Ximing Lu, Youngjae Yu, Yanpeng Zhao, Mohammadreza
  Salehi, Aditya Kusupati, Jack Hessel, Ali Farhadi, and Yejin Choi.
\newblock Merlot reserve: Neural script knowledge through vision and language
  and sound.
\newblock In \emph{CVPR}, pp.\  16375--16387, 2022.

\bibitem[Zhang et~al.(2018)Zhang, Hu, and Sha]{zhang2018cross}
Bowen Zhang, Hexiang Hu, and Fei Sha.
\newblock Cross-modal and hierarchical modeling of video and text.
\newblock In \emph{ECCV}, pp.\  374--390, 2018.

\bibitem[Zhang et~al.(2022)Zhang, Guo, Zhang, Li, Miao, Cui, Qiao, Gao, and
  Li]{zhang2022pointclip}
Renrui Zhang, Ziyu Guo, Wei Zhang, Kunchang Li, Xupeng Miao, Bin Cui, Yu~Qiao,
  Peng Gao, and Hongsheng Li.
\newblock Pointclip: Point cloud understanding by clip.
\newblock In \emph{CVPR}, pp.\  8552--8562, 2022.

\bibitem[Zhao et~al.(2022)Zhao, Zhu, Wang, and Yang]{zhao2022centerclip}
Shuai Zhao, Linchao Zhu, Xiaohan Wang, and Yi~Yang.
\newblock Centerclip: Token clustering for efficient text-video retrieval.
\newblock \emph{arXiv preprint arXiv:2205.00823}, 2022.

\bibitem[Zhu \& Yang(2020)Zhu and Yang]{zhu2020actbert}
Linchao Zhu and Yi~Yang.
\newblock Actbert: Learning global-local video-text representations.
\newblock In \emph{CVPR}, pp.\  8746--8755, 2020.

\end{thebibliography}
\bibliographystyle{iclr2023_conference}

\appendix
\section{More Downstream Tasks}
We add various suitable downstream tasks to verify that our model can be generalizable to other tasks. Specifically, we add four zero-shot evaluations on video action recognition and video-language multi-choice tasks:

\paragraph{HMDB51 Action Recognition} HMDB51 \citep{kuehne2011hmdb} is a video dataset containing realistic videos from various sources, including movies and web videos. There are 51 action categories (such as “jump”, “kiss” and “laugh”). We evaluate our model by zero-shot classification on HMDB51. To adapt to our model, we formulate video classification as video-text alignment, using text prompts "a person [label]". The prediction is made on top-1 similarity between query video and prompted texts. We directly test our model without finetuning to report zero-shot accuracy results.

\paragraph{SomethingSomethingV2 (SSv2) Action Recognition.} SomethingSomethingV2 \citep{goyal2017something} dataset is a large collection of labeled video clips that show humans performing pre-defined basic actions with everyday objects. It requires models to develop fine-grained understanding of basic actions that occur in the physical world. To adapt to our model, we design a multi-choice task on SSv2. The model has to choose the ground truth annotation among all categories. We directly test our model without finetuning to report zero-shot accuracy results. 

\paragraph{MSR-VTT Multi-Choice (MC).} MSR-VTT MC is a benchmark build on MSR-VTT video dataset. Given a video query and five descriptive sentences, MSR-VTT MC requires the model to choose a single best answer in the candidates. We directly test our model without finetuning to report zero-shot accuracy results.

\paragraph{LSMDC Multi-Choice (MC).} LSMDC MC is a benchmark build on movie data. Similar with MSR-VTT MC, it requires the model to choose a single best answer from 5 candidates. We report zero-shot accuracy on 10000 test videos.

\begin{table*}[!h]
    \centering
    \begin{tabular}{l c c c c} 
    \toprule
    Model & HMDB51 & SSv2 & MSR-VTT MC & LSMDC MC  \\
    \midrule
    CLIP-MeanPool & 40.3 & 25.8 & 89.9 & 64.3 \\
    CLIP-ViP w/o post-pretraining & 41.1 &	26.1 &	89.4 &	63.5 \\
    CLIP-ViP w/ video-subtitle &	36.6 &	28.0 &	86.6 &	67.0 \\
    CLIP-ViP full &	\bf44.8 &\bf	33.0 &	\bf90.3 &	\bf67.6 \\
    \bottomrule
    \end{tabular}
    \caption{Evaluation on more downstream tasks. We report zero-shot accuracy (\%) on HMDB51, SSv2, MSRVTT-MC, and LSMDC MC. All models are base size with patch size of 32. Details of post-pretraining are the same as in Section \ref{sec:abl}.}
    \label{tab:more}
\end{table*}

From the results in Table \ref{tab:more}, we can find that post pretraining CLIP on subtitle alone will lead to a performance drop on most tasks: HMDB51 SSv2 and MSR-VTT MC. This phenomenon is in consistency with our assumption as subtitles usually have unique form. By our final setting, CLIP-ViP achieves better results than CLIP on all evaluations, especially on SSv2 which requires more understanding on temporality \citep{wang2022bevt}. These results demonstrate that CLIP-ViP outperforms CLIP on various video benchmarks and further show the generalization of our model CLIP-ViP.

\section{Efficiency of Post-pretraining}
In light of the good image representation of CLIP, our post-pretraining significantly reduces the training cost, Compared with HD-VILA \citep{xue2022advancing} which trains video-language representation from scratch, CLIP-ViP outperforms HD-VILA by a large margin with about 1/30 training time. To make more fair comparison with training from scratch. We conduct an experiment of training from scratch with the same setting and training time (one epoch). Then we finetune on MSR-VTT Retrieval to evaluate the pre-trained model.

\begin{table*}[!h]
    \centering
    \begin{tabular}{l c c c} 
    \toprule
    Model & MSR-VTT R@1 &	MSR-VTT R@5 &	MSR-VTT R@10  \\
    \midrule
    CLIP-ViP (from scratch)	& 28.5 & 55.2 &	68.6 \\
    CLIP-ViP &	49.6 &	74.5 &	84.8 \\
    \bottomrule
    \end{tabular}
    \caption{Comparison with training CLIP-ViP from scratch. We report finetuning results on MSR-VTT Retrieval. All models are base size with patch size of 32. Details of post-pretraining are the same as in Section \ref{sec:abl}.}
    \label{tab:scratch}
\end{table*}

From the results in Table \ref{tab:scratch}, we can see that there is a huge performance gap. Without basing on a CLIP, CLIP-VIP can not leverage the rich knowledge entailed in CLIP, thus leading to a much lower training efficiency.

\section{Responsible AI Considerations}
The proposed video-language pre-training model shows the capacity and generalization of learned VL representation which could benefit many applications of CV and NLP with a large range of uses across many domains. Each one of the uses has potential benefits and societal impacts. While we foresee that our technology could be used to find key information and improve efficiency and effectiveness for helpdesks, recommendation, retail and sales, we realize that it could also be used, in combination with new data, to fine-tune models to mislead, or otherwise harm people. We are also aware that this work uses considerable computation resources which itself, has environmental impacts. Therefore reducing the model size and computing effort is essential for future research.

Machine learning systems can display unfair behavior for different individuals or groups. This is a multi-dimensional, socio-technical challenge and is not explicitly addressed or captured in the current accuracy metrics for this research technology. In general, standardized fairness measures have not yet been agreed upon in academia or industry. We see opportunities for more work in this area to develop methods and benchmarks for measuring fairness aspects.

Given that user generated data is used, it is possible that certain demographic groups may not have enough representation. While we have balanced various video categories to mitigate for disparities, it is still likely that bias and fairness issues exist; this is an area of potential future work.  There may be a Western heteronormative bias, stereotypical depictions of historically marginalized populations and/or lack of representation among some groups. Although we have filtered the input data for explicit and violent content, it is possible that it hasn’t been totally eliminated in the training data and could have impacts on the results.

While some mitigations for potential harms can be done in the base model, it’s important to recognize that considering risks for fine-tuning data for particular scenarios is critical as well. Ultimately, choosing the application scenario of any final model used in a production system will require careful consideration of potential harms specific to the scenario. 


\end{document}